\documentclass[accepted]{melba}

\usepackage{mwe} 

\usepackage{amsmath,amsfonts}

\usepackage{array}
\usepackage{adjustbox}

\DeclareMathOperator*{\E}{\mathbb{E}}

\melbaid{2025:028}  
\doi{10.59275/j.melba.2025-gb33}
\melbaauthors{Sun, Woerner, Maier, Koch, Baumgartner}
\email{susu.sun@uni-tuebingen.de}
\volume{3}
\firstpageno{636}  
\melbayear{2025}  
\datesubmitted{2025-01-06}  
\datepublished{2025-10-20}  

\melbaspecialissue{Medical Imaging with Deep Learning (MIDL) 2020}
\melbaspecialissueeditors{Marleen de Bruijne, Tal Arbel, Ismail Ben Ayed, Hervé Lombaert}

\ShortHeadings{Attri-Net: A Globally and Locally Inherently Interpretable Model}{Sun, Woerner, Maier, Koch, Baumgartner}

\title{Attri-Net: A Globally and Locally Inherently Interpretable Model for Multi-Label Classification Using Class-Specific Counterfactuals}

\author{
        \firstname Susu \surname Sun\aff{1},\orcid{0009-0009-8635-9164}
        \firstname Stefano \surname Woerner\aff{1},
        \firstname Andreas \surname Maier\aff{2}, \orcid{0000-0002-9550-5284}
        \firstname Lisa M. \surname Koch\aff{3,4},\orcid{0000-0003-4377-7074}
        \firstname Christian F. \surname Baumgartner\aff{1,5},\orcid{0000-0002-3629-4384}
}
\affiliations{
        \num 1 \addr Cluster of Excellence: Machine Learning - New Perspectives for Science, University of Tübingen, Tübingen, Germany \\
        \num 2 \addr Department of Computer Science, Friedrich-Alexander-Universität Erlangen-Nürnberg, Erlangen, Germany \\ 
        \num 3 \addr Hertie Institute for Artificial Intelligence in Brain Health, University of Tübingen, Tübingen, Germany \\
        \num 4 \addr Department of Diabetes, Endocrinology, Nutritional Medicine and Metabolism, Inselspital, Bern University Hospital, University of Bern, Bern, Switzerland \\
        \num 5 \addr Faculty of Health Sciences and Medicine, University of Lucerne, Lucerne, Switzerland\\
}

\abstract{
	Interpretability is crucial for machine learning algorithms in high-stakes medical applications. However, high-performing neural networks typically cannot explain their predictions. Post-hoc explanation methods provide a way to understand neural networks but have been shown to suffer from conceptual problems. Moreover, current research largely focuses on providing local explanations for individual samples rather than global explanations for the model itself. In this paper, we propose Attri-Net, an inherently interpretable model for multi-label classification that provides both local and global explanations. Attri-Net first counterfactually generates class-specific attribution maps to highlight the disease evidence, then performs classification with logistic regression classifiers based solely on the attribution maps. Local explanations for each prediction can be obtained by interpreting the attribution maps weighted by the classifiers' weights. Global explanation of whole model can be obtained by jointly considering learned average representations of the attribution maps for each class (called the class centers) and the weights of the linear classifiers. To ensure the model is ``right for the right reason", we introduce a mechanism to guide the model's explanations to align with human knowledge. Our comprehensive evaluations show that Attri-Net can generate high-quality explanations consistent with clinical knowledge while not sacrificing classification performance. 
	Our code is available at~\url{https://github.com/ss-sun/Attri-Net-V2}.}

\begin{document}

\twocolumn[\maketitle]

\begin{figure*}[t]
\centerline{\includegraphics[width=0.8\linewidth]{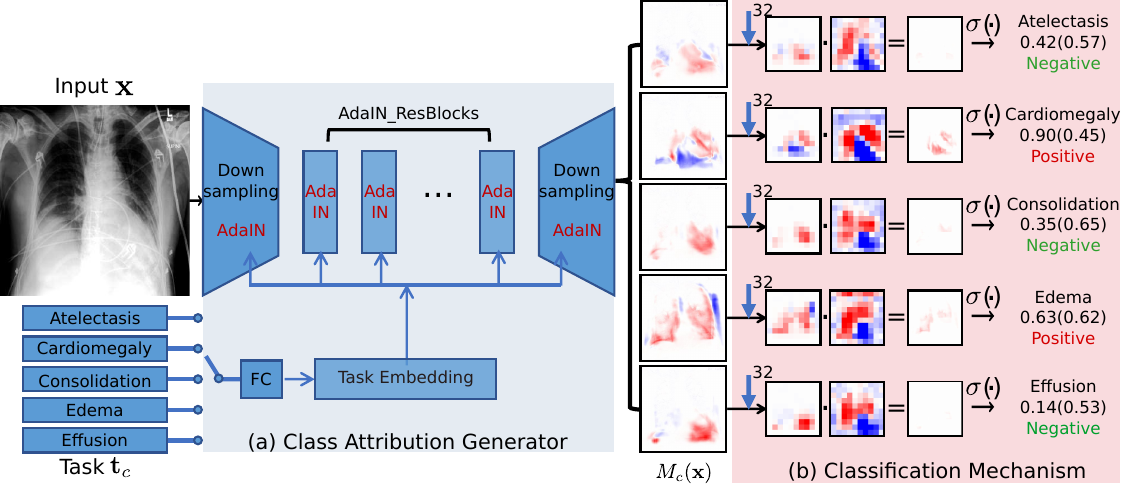}}
\caption{Overview of the Attri-Net framework. Given an input image $\textbf{x}$ and a diagnostic task $\mathbf{t_c}$, the visual feature attribution generator \textbf{(a)} produces counterfactual attribution maps $M_c(\mathbf{x})$ that highlight specific disease effects. Logistic regression classifiers in \textbf{(b)} produce the final prediction for each class based on downsampled versions of these attribution maps.}
\label{fig:framework}
\end{figure*}

\section{Introduction}
    
\enluminure{D}{eep} neural networks have significantly improved the performance of various medical image analysis tasks in experimental settings \citep{litjens2017survey}. However, the black-box nature of deep learning models may lead to a lack of trust \citep{dietvorst2015algorithm}, or more concerningly, blind trust among clinicians \citep{tschandl2020human, gaube2021ai}. Using black-box models may potentially also result in ethical and legal problems \citep{grote2020ethics}, thus hindering their clinical adoption. 

Interpretability has been identified as a crucial property for deploying machine learning technology in high-stakes applications such as medicine \citep{rudin2019stop}. The majority of explanation techniques fall into the category of \textit{post-hoc} methods, which are model-agnostic and can generate explanations for any pre-trained model. However, post-hoc methods are not guaranteed to be faithful to the model's true decision mechanism \citep{adebayo2018sanity,arun2021assessing}. Furthermore, as we show in this paper, current post-hoc explanation techniques that were developed for multi-class natural image classification do not perform adequately in the multi-label scenario, in which multiple medical findings may co-occur in a single image. 

An alternative way for approaching interpretability is to design inherently interpretable models, where the explanations are directly built into the model architecture. Recent works such as prototype-based neural networks \citep{chen2019looks}, concept bottleneck models \citep{koh2020concept} or B-cos Networks~\citep{bohle2024b} can provide explanations for their predictions by revealing the actual decision process to the user. Thus, these explanations are considered to be faithful to the model's internal mechanism.

Based on the scope, the explanations can also be categorized as local or global explanations. Local explanations focus on understanding specific predictions for individual samples, while the global explanations aim to understand the overall decision mechanism of the entire model on the dataset level \citep{christoph2020interpretable, bassan2024local}. Although prior research has primarily focused on local explanations \citep{bassan2024local}, for medical applications, it is also important to understand the behavior of machine learning models on a global level. Global explanations offer a valuable tool for verifying learned features and identifying potential spurious correlations, such as the model's reliance on task-irrelevant information. This is especially important since deep neural networks have a tendency to exploit shortcuts rather than focusing on task-relevant features \citep{geirhos2020shortcut}. This shortcut learning behavior, as highlighted in \citep{sagawa2019distributionally}, can significantly undermine the model's robustness and generalization capabilities. 

To encourage deep neural networks to focus on task-relevant features for making predictions and increase the generalization of the model, recent research \citep{erion2021improving, pillai2022consistent, rao2023studying, li2018thoracic} has explored diverse methods of guiding models, such as enforcing desirable properties on the attributions or aligning explanations with human annotation. Given that many explanation methods are differentiable \citep{selvaraju2017grad, bohle2024b, shrikumar2017learning}, model guidance can be directly integrated with these explanations, allowing optimization for both classification performance and feature localization.

In this work, we propose Attri-Net, an inherently interpretable model designed for the multi-label classification scenario. The key contributions of this work are as follows: Attri-Net provides faithful local explanations for individual predictions and global explanations that reveal the entire model's behavior at the dataset level. We incorporate a model guidance mechanism into Attri-Net to encourage the model to be ``right for the right reason", relying on minimal pixel-wise disease annotations, which is typical given the scarcity of expert annotations in the medical domain. Quantitative and qualitative evaluations show that Attri-Net generates high-quality local explanations while retaining classification performance comparable to state-of-the-art models. Furthermore, we demonstrate that Attri-Net's global explanation can effectively identify the spurious correlation learned by the model in short-cut learning settings, and the proposed model guidance mechanism can successfully mitigate this undesired behavior.

This work extends our previous conference paper \citep{sun2023inherently} by introducing the global explanation that captures the model’s overall behaviors at the dataset level. We design new experiments to demonstrate its effectiveness in identifying shortcut learning tendencies. Furthermore, we incorporate a model guidance mechanism that leverages human knowledge to guide the model during the training process, encouraging the model to make decisions that are ``right for the right reason". New experiments show that this model guidance mechanism can effectively mitigate shortcut learning behavior. In addition, we expand the literature review and provide a more in-depth discussion of the limitations of our current approach and directions for future work.

\section{Related Works}


\subsection{Post-hoc explanation methods}

Post-hoc explanation methods are model agnostic and can be used to explain any trained model. A widely used group of techniques in this category are gradient-based methods such as Guided Backpropagation \citep{springenberg2014striving} or Integrated Gradients \citep{sundararajan2017axiomatic}. Those techniques provide insights into black-box neural networks by visualizing the gradient with respect to the input pixels. Recent studies \citep{adebayo2018sanity,arun2021assessing} have demonstrated that these explanations do not change significantly when the target model changes, raising concerns about the ability of such post-hoc methods to faithfully reflect the model's behavior.

Perturbation-based methods such as LIME \citep{ribeiro2016should} and SHAP \citep{lundberg2017unified} approximate the decision function by perturbing the input and observing the changes in the output. Also for these methods the explanation's faithfulness to the model's decision mechanism is not guaranteed. Due to their reliance on input perturbations, these methods can exhibit high variability across different runs. Factors such as how the input is segmented into parts~\citep{pihlgren2025segmentationsmoothingaffectexplanation}, the sampling procedures~\citep{zhang2019should}, and the methods used to assign importance to the parts~\citep{chen2018shapley} all contribute to this variance. As a result, the explanations produced by these methods are often not robust.

Another line of work, including Class Activation Mappings (CAM)~\citep{zhou2016learning} and GradCAM~\citep{selvaraju2017grad} are based on the networks' final activation map and highlight the regions that are important for a specific class. These techniques are limited by the spatial resolution of their explanations. Moreover, Grad-CAM may highlight regions of an image that a model did not actually use for prediction \citep{draelos2020use}.

\subsection{Inherently interpretable models} 

In contrast to post-hoc explanations, some recent work has focused on designing models that are inherently interpretable. These models are constructed such that the explanations are a built-in part of the decision mechanism. Therefore, the explanations can faithfully reveal the true decision mechanism. 

Examples of inherently interpretable methods include rule-based models \citep{lakkaraju2016interpretable, angelino2018learning} which break the decision mechanism into a set of independent if-then rules that are easily interpretable by humans. Inherently interpretable concept bottleneck models \citep{alvarez2018towards,chen2020concept,koh2020concept} first make predictions on human-interpretable concepts, and then use these concept activations to generate final predictions. This structure provides human-understandable explanations and enables human intervention. Prototype-based inherently interpretable models \citep{chen2019looks,barnett2021interpretable} learn a set of prototypes from training images and make predictions by comparing regions of the input image to these prototypes. Both the final prediction and the explanation are derived from the similarity to the learned prototypes. Concept-based methods and prototype-based methods both provide human-friendly explanations.
However, these methods do not provide spatial explanations, typically involve many hyperparameters and complex training regimes, and are challenging to train.

A number of works aim to generate inherently interpretable spatial explanations. For instance, BagNets \citep{brendel2019approximating, donteu2023sparse} produce spatial explanation based on small local image patches that contain class evidence. Recently proposed models like CoDA-Net \citep{bohle2021convolutional} and its more generalized version, B-cos Networks \citep{bohle2024b} are currently the state-of-the-art models for providing inherently interpretable visual explanations at the pixel level. These methods employ a dynamic alignment mechanism and formulate the networks' prediction as a weighted sum of the input images. 

\subsection{Counterfactual-based explanations}
Counterfactual-based explanations are a category of methods highly relevant to our work. These methods attempt to answer questions like ``What would the image look like if it belonged to a different class?" \citep{schutte2021using,joshi2018xgems,boreiko2022visual, jeanneret2023adversarial}, or they aim to exaggerate features pertinent to the predicted class \citep{cohen2021gifsplanation,singla2019explanation}. 

Generative adversarial networks (GAN) \citep{goodfellow2020generative} are widely used for generating counterfactual explanations. For example, \cite{atad2022chexplaining} generate counterfactual explanations for Chest X-rays by manipulating the latent style space of StyleGAN. \cite{mindlin2023abc} employ CycleGAN to produce attention-based counterfactuals for X-ray image classification. \cite{garg2024advancing} propose a GAN-based ante-hoc explainable classifier. And \cite{qi2025projectedex} leverage the style space of StyleGAN to generate counterfactual explanations for classifier decisions on prostate MRI scans.

Apart from GANs, Autoencoders \citep{bank2023autoencoders} have also been explored for generating counterfactual explanations. The most relevant one with our work is the Gifsplanation \citep{cohen2021gifsplanation, cohen2025identifyingspuriouscorrelationsusing}, which generates counterfactual explanations for chest x-rays by shifting the latent space of an autoencoder. 

Recently, the Denoising Diffusion Probabilistic Models (DDPM) \citep{ho2020denoising} and its variants have achieved remarkable success in the generative tasks and have been explored for generating counterfactual explanations \citep{augustin2022diffusion}. For example, \cite{jeanneret2023adversarial} generate post-hoc counterfactual explanations using DDPM, 
\cite{bedel2024dreamr} propose DreamMR, which is the first diffusion-driven counterfactual explanation method for functional MRI. And \cite{fathi2024decodex} 
propose DeCoDEx to improve Diffusion-based Counterfactual Explanations in confounder detection.

Although there are various counterfactual-based explanation approaches, all models that we are aware of generate explanations in a post-hoc manner, which may lead to unfaithful explanations. To our knowledge, we present the first inherently interpretable model that leverages counterfactuals.

\subsection{Global interpretability}
One important goal of interpretability is to detect and avoid bias of the ML models \citep{ghassemi2021false}. In contrast to local explanations, which reveal the decision mechanism for individual samples, global explanations provide insights into the model behavior for an entire dataset \citep{reyes2020interpretability} and are therefore particularly helpful for detecting unwanted model behavior. The vast majority of prior research has focused on local explanations, with very few techniques tackling the global explanation problem. Post-hoc explanation SHAP can provide global explanations by running it on every sample and aggregating the SHAP values across the entire dataset, thereby offering insights into the model as a whole. The prototypes in prototypical networks \citep{snell2017prototypical, chen2019looks} represent the cluster centers of each class and can serve as a global explanation for the classifier. \cite{kim2016examples} proposed a technique capable of global explanations by learning both representative prototypical examples of the dataset and criticism examples that do not quite fit the model. Concept activation vectors are another strategy that allows obtaining explanations of entire classes or sets of examples \citep{kim2018interpretability}. The system proposed in \citep{pereira2018enhancing} achieved both local and global interpretability by jointly considering existing correlations between imaging data, features, and target variables. To our knowledge, our work is the first inherently interpretable model that provides both local and global faithful explanations.

\subsection{Visual feature attribution} 

Visual feature attribution is a task closely related to building explainable deep learning models. Rather than obtaining insights into a model's decision mechanism, visual feature attribution methods aim to visualize evidence of a particular class in an image. We emphasize that visual feature attribution is distinct from interpretable machine learning because its aim is not to predict an outcome. Nevertheless, the most frequently used approach to address this problem is training a neural network for the classification task and then employing spatial post-hoc explanation techniques described earlier~\citep{baumgartner2017sononet, jamaludin2016spinenet,zhu2017deep, pinheiro2015image}. \cite{baumgartner2018visual} pointed out that these algorithms may lead to only a subset of class-relevant features being detected since not all class-relevant pixels are necessarily used by a classifier for its prediction. The authors proposed an alternative strategy for visual feature attribution based on counterfactuals produced using a Wasserstein GAN. Specifically, the residual between the original image and the generated counterfactual image of the opposite class was used to identify class relevant features. However, the technique requires knowledge of the ground-truth label \emph{a priori} and can therefore not be used for classification. Moreover, the technique is limited to the binary scenario with a healthy and a pathological class. In our work, we build on this work to develop an interpretable multi-label classifier based on counterfactuals.

\subsection{Model guidance}

Deep neural networks make predictions by recognizing the discriminative features learned from images in the training set. However, deep neural networks have a tendency to exploit shortcuts, and the learned features may not necessarily transfer to the unseen images \citep{geirhos2020shortcut}. To enhance model generalization and reduce potential bias, recent research has focused on incorporating task-relevant information into model guidance. For instance, \cite{fathi2024decodex} introduced a framework that employs a pre-trained spurious correlation detector to improve the accuracy of diffusion-based counterfactual explanations. \cite{erion2021improving} introduced attribution priors such as smoothness and sparsity to the model during training to optimize for higher-level properties of explanations.
\cite{pillai2022consistent} proposed Contrastive Grad-CAM Consistency to regularize the model to produce more consistent explanations. Prior studies \citep{li2018thoracic, rao2023studying} have demonstrated the effectiveness of bounding box annotations in guiding the model. And \cite{rao2023studying} has extensively evaluated various aspects, including loss functions, attribution methods, and depth of model guidance, concluding that incorporating human knowledge guidance into the model can enhance the interpretability of the explanations and mitigate potential shortcut learning behavior. In this work, we investigate guiding our model through pixel-wise human annotations. We use the energy loss proposed in \citep{rao2023studying} to encourage our model to be ``right for the right reason".

\section{Methods}\label{sec:methods}
\subsection{Framework Overview}\label{sec:framework_overview}

In this section, we introduce our proposed inherently interpretable ``Attri-Net" model. Attri-Net is a multi-label classifier that makes predictions for $C$ classes, where each class $c$ with label $y_c \in \{0,1\}$ corresponds to the presence or absence of a specific medical finding in an image. 

The core idea of our approach is to first counterfactually generate class attribution maps containing the evidence for each diagnostic task $\mathbf{t}_c$ (Fig.~\ref{fig:framework}(a)), and then
perform classification solely based on the attribution maps using logistic regression classifiers (Fig.~\ref{fig:framework}(b)). The attribution maps weighted by the logistic regression weights directly show how specific predictions are calculated and provide local explanations. A global explanation can be obtained by jointly visualising learned average representations of the attribution maps for each class (class centers) and the classfiers' linear weights. 

\subsection{Class Attribution Generator}

\begin{figure}[h]
    \centering
    \includegraphics[width=1\linewidth]{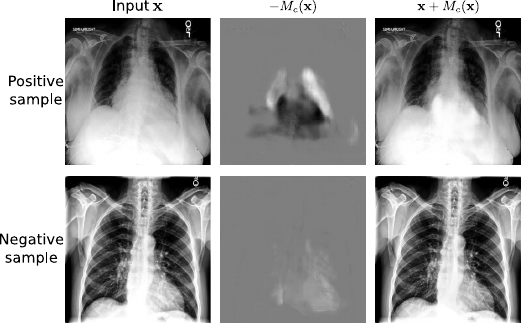}
    \caption{Examples of counterfactual images. The top row shows an input image $\mathbf{x}$ that is positive in cardiomegaly, the attribution map $M_c(\mathbf{x})$ (with a flipped sign for better visualization effect), and the counterfactual image $\hat{\mathbf{x}}$. The bottom row shows images for a negative sample. As expected the residual changes $M_c(\mathbf{x})$ are large for the positive sample and small for the negative sample.}
    \label{fig:counterfactual}
\end{figure}

\label{sec:class_attribution_generator}
The central component of Attri-Net is the class attribution generator (see Fig.~\ref{fig:framework}(a)). We define this generator as an image-to-image translation network, denoted as $M_c(\mathbf{x}): \mathbb{R}^{H \times W} \mapsto \mathbb{R}^{H \times W}$, following the approach in \citep{baumgartner2018visual}. It learns an additive mapping $M_c$ to produce a counterfactual image $\hat{\mathbf{x}}$, defined as 
\begin{equation*}
    \hat{\mathbf{x}} = \mathbf{x} + M_c(\mathbf{x}) \text{\,.}
\end{equation*}
The objective is to make the generated counterfactual image $\hat{\mathbf{x}}$ indistinguishable from the real images sampled from the distribution $p(\mathbf{x}|y_c=0)$ that do \textit{not} contain class $c$.

Intuitively, $M_c$ captures the changes required for each pixel in the input to remove the positive effect of class $c$ from the image. Consequently,  $M_c$ will contain larger changes for images from the positive class $p(\mathbf{x}|y_c=1)$ compared with images from the negative class $p(\mathbf{x}|y_c=0)$. An illustrative example of generated counterfactuals $\hat{\mathbf{x}}$ for the disease ``cardiomegaly" is presented in Fig.~\ref{fig:counterfactual}, with additional examples available in the Appendix \ref{sec:examples_counterfactual}.

In order to enable $M_c$ to learn the difference between the class-positive and class-negative distributions of class $c$, we employ a class-specific discriminator network $D_c$. The network $D_c$ is trained alongside $M_c$ to distinguish fake images ($\mathbf{x} + M_c(\mathbf{x})$) from real negative class images from $p(\mathbf{x}|y_c=0)$. Specifically, we adopt the adversarial Wasserstein GAN loss~\citep{arjovsky2017wasserstein,baumgartner2018visual}. The discriminator loss can be written as

\begin{align}
    \mathcal{L}^{(c)}_{\text{disc}} &= \E_{p(\mathbf{x}|y_c=0)} [-D_c(\mathbf{x})] \nonumber \\
    &+ \E_{p(\mathbf{x}|y_c=1)} [D_c(\mathbf{x} + M_c(\mathbf{x})) ] \text{\,.}
\end{align}

The adversarial class attribution generator loss maximises the second term of the equation above: 
\begin{equation}
    \mathcal{L}^{(c)}_{\text{adv}} = \E_{p(\mathbf{x}|y_c=1)} [-D_c(\mathbf{x} + M_c(\mathbf{x})) ] \text{\,.}
\end{equation}
The class attribution generator loss ensures that $\hat{\mathbf{x}}$ is a realistic counterfactual not containing class $c$ and, by extension, that $M_c$ is a realistic residual attribution map containing all positive evidence of class $c$.

To encourage the class attribution maps to focus on class-relevant information and avoid learning superfluous pixels not belonging to a given class, we incorporate an additional $L_1$ regularization term on $M_c$ to encourage it to be sparse. The regularization term is defined as follows:

\begin{align}
\mathcal{L}^{(c)}_{\text{reg}} &= \alpha_0 \E_{p(\mathbf{x}|y_c=0)} [{\lVert M_c(\mathbf{x}) \rVert }_1] \nonumber \\
&+ \alpha_1 \E_{p(\mathbf{x}|y_c=1)} [ {\lVert M_c(\mathbf{x}) \rVert }_1 ] \text{\,.}
\end{align}

We assign a higher weight, denoted as $\alpha_0$, to samples from the class-negative category and a lower weight, denoted as $\alpha_1$, to class-positive examples. This weighting reflects the intuition that minimal adjustments are required for samples from the negative class compared to those from the positive class. We choose $\alpha_0 = 2, \alpha_1 = 1$ based on preliminary parameter tuning experiments and use them for all experiments in this paper.

In the context of the multi-label classification task, our objective is to generate individual explanations for each medical finding. Although it is feasible to design a network $M$ to produce class attribution maps for all classes as multiple output channels in a single forward pass, preliminary experiments on such an architecture revealed inadequate class attribution in the multi-label scenario. Instead, we introduce a task switch mechanism based on recent work~\citep{sun2021task} to enable the class attribution generator to switch between various diagnostic tasks. As shown in Fig.~\ref{fig:framework}(a), the task code $\mathbf{t}_c$ is injected into the class attribution generator through adaptive instance normalization (AdaIN) layers to switch the network $M$ to a specific mode $M_c$ that focuses on diagnosing class $c$. Each task code $\mathbf{t}_c$ is a one-hot encoding spatially upsampled by a factor of 20 as in~\citep{sun2021task}. Then the one-hot vector task code is converted into a task embedding via a small fully connected network and fed to AdaIN layers which are placed throughout the network. We apply the same mechanism also to the discriminator network $D$ such that it can provide correct feedback to the respective class.

With task switching mechanism, the class attribution generator and discriminator can now be expressed as ~$M_c(\mathbf{x}) \\ = M(\mathbf{x}, \mathbf{t}_c)$, and $D_c(\mathbf{x}) = D(\mathbf{x}, \mathbf{t}_c)$. The class attribution maps for all labels can be obtained by repeated forward passes through $M$ while iterating through the $\mathbf{t}_c$ vectors of all classes. The specific architecture of $M$ and $D$ is discussed in Sec.\ref{sec:model_architecture_and_training}, and in more detail in Appendix \ref{sec:training}.

\subsection{Classification Mechanism}
\label{sec:classification_layers}
As $M(\mathbf{x}, \mathbf{t}_c)$ learns the changes required to convert an image $\mathbf{x}$ into a sample from the negative distribution $p(\mathbf{x}|y_c=0)$, it inherently encodes class-specific information and can be directly used for predicting class $c$. Test images that contain the disease will require large changes to make them appear healthy, while images that are already healthy are characterised by very small changes $M(\mathbf{x}, \mathbf{t}_c)$ (see Fig.~\ref{fig:counterfactual}). Since the maps $M(\mathbf{x}, \mathbf{t}_c)$ allow easy differentiation of positive and negative samples for a given disease, we can employ a simple linear classifier for the final classification step of each class. As shown in Fig.~\ref{fig:framework}(b), the respective attribution map is downsampled and then used as input to the classifier. That is, 
\begin{equation}
\label{eq:logreg}
p(y_c|\mathbf{x}) = \sigma\Bigl(\sum_{i,j} w_{ij}^{(c)} \cdot S_\gamma (M(\mathbf{x}, \mathbf{t}_c))_{ij}\Bigr) \text{\,,}
\end{equation}
where $S_\gamma$ is a 2D average pooling operator that downsamples by a factor of $\gamma$, $w_{ij}^{(c)}$ denotes the weights associated with each pixel of the down-sampled attribution map for class $c$, and $\sigma$ is the sigmoid function. In preliminary experiments, we evaluated with various values of $\gamma$ and found $\gamma=32$ to perform robustly and we used this value for all experiments. 

The classifiers for each class are trained using a standard binary classification loss $\mathcal{L}^{(c)}_{\text{cls}}$, i.e. a binary cross entropy loss. Note that, since our framework is trained end-to-end, $M$ also receives gradients from that loss and is thereby encouraged to create class attribution maps that are linearly classifiable.

\subsection{Center Loss}
\label{sec:center_loss}

To further encourage the attribution maps to be discriminative, we apply the center loss proposed by \citep{wen2016discriminative}, which has demonstrated its efficacy in fostering more distinctive feature representations. Extending this idea, we define center loss $\mathcal{L}^{(c)}_{\text{ctr}}$ as follows:

\begin{align}
    \mathcal{L}^{(c)}_{\text{ctr}} &= \frac{1}{2} 
\E_{p(\mathbf{x}|y_c=0)} \left[ {\lVert M(\mathbf{x}, \mathbf{t}_c) - \mathbf{v}_{y_c=0} \rVert }_2^2 \right] \nonumber\\
    &+ \frac{1}{2} \E_{ p(\mathbf{x}|y_c=1)} \left[ {\lVert M(\mathbf{x}, \mathbf{t}_c) - \mathbf{v}_{y_c=1} \rVert }_2^2 \right] \text{\,,}
\end{align}
where $\mathbf{v}_{y_c=0}, \mathbf{v}_{y_c=1} \in \mathbb{R}^{H \times W}$ are the negative and positive class centers of attribution map for class $c$. 

The class centers are learnable and updated on mini-batch along with the update step of network $M$ with separate optimizers as described by \citep{wen2016discriminative}. In the forward pass, the center loss calculates the $L_2$ distance between the attribution map $M(\mathbf{x},\mathbf{t}_c)$ and the corresponding class center. In the backward pass, both the attribution map $M(\mathbf{x},\mathbf{t}_c)$ and the associated class center receive gradients from the loss and are updated accordingly. Since the center loss penalizes the distance between an attribution map and its corresponding class center, it encourages the attribution map to move closer to the class center. This reduces the intra-class distance while increasing the inter-class separation, making the attribution map more distinctive between positive and negative classes. 
Furthermore, the class centers aggregate the mean attribution maps of each class $c$ across the dataset, offering insight into the model's average behavior. At the end of the training stage, these class centers can serve as a global explanation for the entire model as described in Sec.\ref{sec:obtaining_global_explanations}.

\subsection{Model Guidance}
\label{sec:model_guidance_loss}

\begin{figure}[h]
    \centering
    \includegraphics[width=1\linewidth]{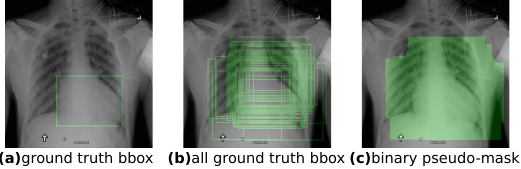}
    \caption{Generation of pseudo guidance masks. An example for the disease cardiomegaly from the ChestX-ray8 dataset is shown. \textbf{(a)} A chest X-ray image with its ground truth bounding box annotation for cardiomegaly. \textbf{(b)} The same image with multiple cardiomegaly bounding box annotations from other cases in the ChestX-ray8 dataset. \textbf{(c)} The same image with the binary pseudo-mask generated from multiple cardiomegaly bounding box annotations.}
\label{fig:pseudo_bbox}
\end{figure}

The inherently interpretable construction of Attri-Net allows to constrain the explanations using human guidance in the form of disease segmentations or bounding boxes. Since the explanations are part of the decision mechanism, constraining the explanations directly affects the model's decision mechanism. 

To incorporate guidance into our model we propose to train it with a guidance loss based on an energy-based formulation~\citep{rao2023studying}:
\begin{equation}
    \mathcal{L}^{(c)}_{\text{guid}} = 1-\frac{\sum_{h=1}^{H} \sum_{w=1}^{W} G_{c_,hw} |M_c|_{,hw}}{\sum_{h=1}^{H} \sum_{w=1}^{W} |M_c|_{,hw}}\text{\,.}
\end{equation}
Here, the guidance mask $G_c$ is a binary image matching the input X-ray's dimensions, where pixels inside lesion regions are set to 1 and others to 0. It is derived from ground-truth lesion annotations, such as bounding boxes or disease segmentations. $|M_c|$ is the absolute value of the attribution map.
The guidance loss encourages the model to focus on the regions within the guidance mask that contain task-relevant features while ignoring the regions outside. 

\subsubsection{Full guidance}\label{sec:full_guidance}

In the ideal case, annotations are available for every training image and we can perform full guidance by directly incorporating the guidance loss with the other loss terms to jointly optimize for classification performance and localization of task-relevant features. This is, for example, the case for the VinDr-CXR dataset~\cite{nguyen2022vindr}, where expert-labeled bounding box annotations are available for every sample. 

\subsubsection{Pseudo-guidance}\label{sec:pseudo_guidance}

In most cases, expert-labeled bounding boxes or segmentations are costly to obtain and are available only for a small portion of the training data. For example, the large-scale chest X-ray dataset ChestX-ray8 \citep{wang2017chestx} contains 108,948 images while only 880 images are provided with disease bounding box annotations. Most of the samples in this dataset only have class labels at the image level. As chest X-ray images primarily capture organs with relatively fixed positions, there is a noticeable overlap of disease incidence regions among individual patients. This intrinsic property of chest X-rays provides an opportunity to incorporate a pseudo guidance mechanism for samples that lack pixel-wise ground truth annotations. We create pseudo-guidance masks for a class by calculating the union of all ground truth bounding box annotations for that class. An example pseudo-mask generation for the class cardiomegaly is shown in Fig.~\ref{fig:pseudo_bbox}.

We investigated different methods to incorporate pseudo-guidance alongside the limited ground truth annotations and determined that a mixed guidance strategy yielded the best performance. Specifically, during training, when a sample has a ground truth annotation, we guide the model using the actual ground truth. In cases where there is no ground truth, we use the pseudo masks as guidance to prompt the model to emphasize regions with a higher prevalence of diseases. We observed that oversampling cases with ground truth annotations during training such that they appear with a frequency of $\tfrac{1}{10}$ enhances the localization performance. We provide a more detailed analysis in the Appendix \ref{sec:ablation_guidance}.

\subsection{Obtaining Explanations}
\label{sec:obtaining_explanations}

During inference time, our model is capable of producing a prediction as well as a local per-sample explanation and a global explanation on the dataset level. In the following we describe how those two explanations types can be obtained. 

\subsubsection{Obtaining Local Explanations}
\label{sec:obtaining_local_explanations}

\begin{figure}[h]
    \centering
    \includegraphics[width=1.0\linewidth]{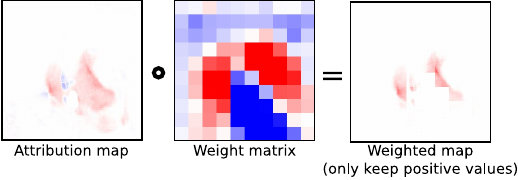}
  \caption{Local explanation of Attri-Net for an example from the CheXpert dataset with cardiomegaly. The weighted attribution map serves as local explanation for a specific prediction. It is defined as the element-wise product of the attribution map from class attribution generator and the weight matrix from corresponding logistic regression classifier.}
  \label{fig:weighted_map}
\end{figure}

Attri-Net directly allows us to obtain local explanations by considering the weighted attribution maps. As shown in Fig.~\ref{fig:weighted_map}, the weighted attribution map is calculated by the element-wise product of the attribution map and the upsampled weight matrix of the corresponding logistic regression classifier. Note that this equivalent to the weights and feature multiplication inside the logistic regression classifiers, i.e. the term inside the sum of Eq.\,\ref{eq:logreg}. The weighted attribution map therefore directly reveals the classifier's decision mechanism. We keep the positive values in the weighted map as evidence of class $c$. This attribution map, together with the final prediction, is provided to the users for inspection of a particular diagnosis.

\subsubsection{Obtaining Global Explanations}
\label{sec:obtaining_global_explanations}

\begin{figure}[h]
    \centering
    \includegraphics[width=1\linewidth]{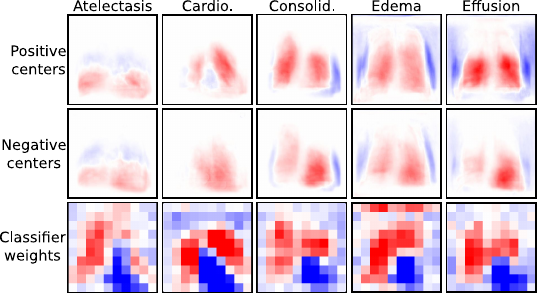}
    \caption{Global explanation of Attri-Net for a model trained on the CheXpert dataset. The positive and negative class centers of attribution maps and the corresponding classifiers’ weight matrix together provide a global explanation for Attri-Net.}
\label{fig:centers_normal}
\end{figure}

Prior to the deployment, model developers need to ensure that the classifier behaves reliably on a global level (e.g. does not use any spurious features). Attri-Net addresses this need by providing global explanations for the whole model's mechanism. 
We define the global explanation as the combination of class centers and classifier weights. After the training stage, the class centers $\mathbf{v}_{y_c=0}, \mathbf{v}_{y_c=1} \in \mathbb{R}^{H \times W}$, as defined in Sec.\ref{sec:center_loss}, capture the average patterns of attribution maps across the entire training set, enabling users to assess whether the input features of the classifiers are clinically meaningful. Meanwhile, the logistic regression classifier weights reveal the areas of the images that the classifier is paying attention to. Thus, the class centers, together with the weight matrices, provide a transparent insight into Attri-Net’s classification process, offering an explanation for the entire model. Figure \ref{fig:centers_normal} illustrates the global explanation for Attri-Net trained on the CheXpert dataset. 

\subsection{Model Architecture and Training}
\label{sec:model_architecture_and_training}

Attri-Net contains a total of 55.86 million parameters and consists of three key components: the class attribution generator $M$, the discriminator network $D$, and the logistic regression classifiers. We implemented $M$ and $D$ based on the StarGAN ~\citep{choi2018stargan} architecture. To enable the task-switching functionality, we replaced instance normalization layers in the original StarGAN architecture with adaptive instance normalization (AdaIN) modules following~\citep{sun2021task}. Each logistic regression classifier was implemented as a fully connected neural network layer with a 2D average pooling layer for downsampling. 
The architectures are described in more detail in the Appendix \ref{sec:training}.

The Attri-Net framework can be trained end-to-end with five loss terms enforcing our essential requirements: Firstly, we used the classification loss $\mathcal{L}_{\text{cls}}^{(c)}$, which apart from encouraging accurate classification, ensures that the attribution maps preserve sufficient class relevant information for a satisfactory classification result. Secondly, we adopted the adversarial loss $\mathcal{L}_{\text{adv}}^{(c)}$ and the regularization term $\mathcal{L}_{\text{reg}}^{(c)}$ to encourage discriminative and sparse attribution maps. Furthermore, the center loss term $\mathcal{L}_{\text{ctr}}^{(c)}$ moves the attribution maps toward the class centers and provides global explanations. Lastly, the guidance loss $\mathcal{L}_{\text{guid}}^{(c)}$ can be optionally used to inject human guidance during training and encourages the attribution maps to be consistent with human knowledge. 

The overall training objective for the class attribution generator $M$ with weight parameters $\varphi$ was given by

\begin{align}
\mathcal{L} &= \sum_c \lambda_{\text{ad}} \mathcal{L}_{\text{adv}}^{(c)}  +\lambda_{\text{cl}} \mathcal{L}_{\text{cls}}^{(c)} +
\lambda_{\text{re}}\mathcal{L}_{\text{reg}}^{(c)} \nonumber \\
&+ \lambda_{\text{ct}} \mathcal{L}_{\text{ctr}}^{(c)} + \lambda_{\text{gd}} \mathcal{L}_{\text{guid}}^{(c)}  \text{\,,}
\end{align}
where we used the hyperparameters $\lambda_*$ to balance the losses. 
We chose $ \lambda_{\text{ad}}=1, \ \lambda_{\text{cl}}=100,  \ \lambda_{\text{re}}=100, \ \lambda_{\text{ct}}=0.01, \ \lambda_{\text{gd}}=30 $ for our experiments. An ablation study on the effect of the different losses can be found in Sec.\ref{sec:ablation_study}.

Throughout the training, we repeatedly iterate through the different classes $c$ and, for each, draw two mini-batches, one containing positive samples of the current class and the other negative samples. We iteratively update $M$, $D$, and the classifiers. In one training step, only the classifier corresponding to the current target class is updated, while the classifiers for other diseases remain frozen. Following the original Wasserstein GAN \citep{arjovsky2017wasserstein}, the discriminator $D$ undergoes more frequent updates to ensure it remains close to optimality throughout training. Specifically, we perform five discriminator update steps for each generator step. Additionally, for every 100 generation step and the first 25 generator steps, we perform an extra 100 discriminator update steps. Furthermore, with each generator step, we update the classifier five times.

We employed the ADAM optimizer~\citep{kingma2014adam} with a learning rate of $10^{-4}$ and a batch size of 4 for the optimization of $M$, $D$, and the logistic regression classifiers. Additionally, following \citep{wen2016discriminative}, we used stochastic gradient descent with a learning rate of 0.1 for updating the class centers in the center loss module. The model was trained for 100,000 generator steps to ensure both the classification performance and the attribution maps achieve a stable state. 
We trained Attri-Net on a single NVIDIA 2080Ti GPU for three days. The best model was chosen by evaluating the area under the ROC curve (AUC) on the validation set. After the training, the optimal decision threshold for each class was obtained by maximising the Youden index (sensitivity + specificity - 1) on the validation set. We also performed this step for the baseline methods. During inference, the average time for predicting all diseases on a single chest X-ray image is 0.194 seconds. Since the attention maps are the intermediate output during inference, generating explanations does not introduce additional computational cost.

\section{Experiments and Results}\label{sec:experiments}
\subsection{Experiment settings}\label{sec:exp_settings}

\subsubsection{Data}

We performed experiments on three chest X-ray datasets: CheXpert \citep{irvin2019chexpert}, ChestX-ray8 \citep{wang2017chestx}, and VinDr-CXR \citep{nguyen2022vindr}. Different from our prior work \citep{sun2023inherently}, we now had access to the officially released test sets of CheXpert and VinDr-CXR datasets. In this paper, we present results for all three datasets based on the official data splits, ensuring the comparability of our findings with other published works that also used the official split. We used the official train, validation and test set for experiments on CheXpert dataset. Since ChestX-ray8 and VinDr-CXR only provide train, test split, we generated train and validation sets on the official train set with a split ratio of 0.8.

We scaled all images to a smaller size of $320 \times 320$ pixels for training and scaled the bounding box annotations and segmentation masks accordingly. Since most of the chest X-ray images in the three datasets are frontal, we excluded a small number of lateral images from the CheXpert dataset. On CheXpert and ChestX-ray8 datasets, following \citep{irvin2019chexpert}, we focused on five findings based on their clinical importance and prevalence: (a) Atelectasis, (b) Cardiomegaly, (c) Consolidation, (d) Edema, and (e) Pleural effusion. The VinDr-CXR dataset is much smaller than the above two datasets and has a different label distribution. To make sure there are enough positive samples for the model to learn disease relevant features, we selected the following five pathologies:
(a) Aortic enlargement, (b) Cardiomegaly, (c) Pulmonary fibrosis, (d) Pleural thickening, and (e) Pleural effusion. 

\subsubsection{Data-splits for Model Guidance}

\begin{figure}[!t]
  \includegraphics[width=1.0\columnwidth]{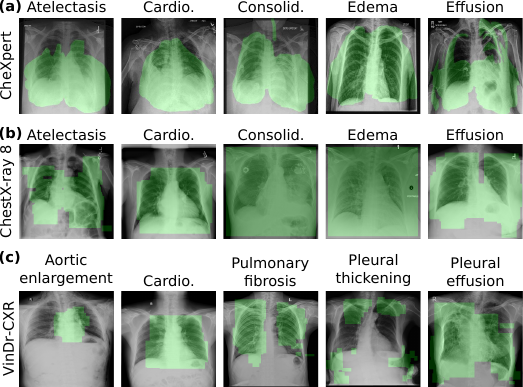}
  \caption{Binary pseudo masks on all three examined datasets.}
  \label{fig:all_pseudo_masks}
\end{figure}

\begin{table}[t]
    \centering
    \caption{Number of samples for creating pseudo guidance.}
    \label{tab:samples_pseudo_guidance}  
    \resizebox{\columnwidth}{!}{
    \begin{tabular}{lccccc}
        \hline
        \bfseries Dataset & \bfseries Atelet. &\bfseries Cardio. &\bfseries Consol.  &\bfseries  Edema &\bfseries Effusi.\\
        \hline
        CheXpert   & 75 & 66 & 32 & 42 & 64  \\
        ChestX-ray8 & 76 & 54 & 0 & 0 & 52\\
  \hline
  \end{tabular}
  }
\end{table}

Though there are a large number of chest X-ray images in the CheXpert and ChestX-ray8 datasets, the number of images with pixel-level annotations of pathologies is much smaller. As explained in Sec.\ref{sec:pseudo_guidance}, we addressed this limitation by generating pseudo masks. These masks were created using the limited ground truth pixel-level annotations and were then used as pseudo-guidance for samples lacking such annotations. In the following we describe how the existing data with pixel-level annotations was split for pseudo-guidance and evaluation.

The ChestX-ray8 dataset provides bounding box annotations for 880 images. We divided these annotations with a ratio of 40\% for generating pseudo masks for training and 60\% for evaluation. Note that in ChestX-ray8, there are no bounding box annotations available for consolidation or edema positive samples. Therefore, we created a very loose guidance in the form of a $300\times300$ square, aiming to guide the model to focus on the central regions of the image. 

In recent work \citep{saporta2022benchmarking}, ground truth segmentations were released for 187 images in the CheXpert validation set and 499 images in the test set. We used the ground truth segmentations from the CheXpert validation set to create pseudo masks, reserving the test set for evaluation. The distribution of samples used to create pseudo guidance can be found in Table \ref{tab:samples_pseudo_guidance}, and the resulting pseudo masks are presented in Fig.~\ref{fig:all_pseudo_masks}.

The VinDr-CXR dataset includes bounding box annotations for all images, allowing us to assess the full guidance we described in Sec.\ref{sec:full_guidance}. As a comparison, we additionally performed an experiment by using the pseudo-guidance mechanism as introduced in Sec.\ref{sec:pseudo_guidance}. We generated pseudo masks by using a small subset of 75 samples from each disease (See Fig.~\ref{fig:all_pseudo_masks}(c)) and only used them to guide the model during training as we did with ChestX-ray8 and CheXpert datasets. This allowed us to evaluate the performance gap between full guidance and pseudo-guidance.

\subsubsection{Baseline methods}

To assess our model's classification performance, we compared it with a standard black-box ResNet50 \citep{he2016deep} model as well as the B-cos ResNet50~\citep{bohle2024b}, a state-of-the-art inherently interpretable classifier. We trained our Attri-Net using the settings described in Sec.\ref{sec:model_architecture_and_training}. For training the ResNet50 and the B-cos ResNet50, we employed the ADAM optimizer~\citep{kingma2014adam} with a learning rate of $10^{-4}$ and a batch size of 4. The models were trained for 50 epochs to achieve convergence, and the best-performing models with the highest AUC score on the validation set were selected. 

We further assessed the local explanations provided by our proposed AttriNet, the B-cos Network, as well as five post-hoc explanation techniques which we applied to the ResNet-50 black-box model. Specifically, we compared to Guided Backpropagation \citep{springenberg2014striving}, GradCAM \citep{selvaraju2017grad}, LIME \citep{ribeiro2016should}, SHAP \citep{lundberg2017unified} and the recently proposed Gifsplanation \citep{cohen2021gifsplanation}. 

Since our model and B-cos Networks are both inherently interpretable, they could both optionally be trained with our guidance loss proposed in Sec.\ref{sec:model_guidance_loss}.  We trained both models with and without model guidance to investigate its effect on explanation quality and classification performance.


\subsection{Evaluation Metrics}
\label{sec:metrics}

\subsubsection{Classification Metric}
\label{sec:classification-metrics}

The classification performance was evaluated using the area under the ROC curve (AUC).

\subsubsection{Explainability Metrics}
\label{sec:explainability-metrics}

\begin{figure}[t]
  \includegraphics[width=1.0\columnwidth]{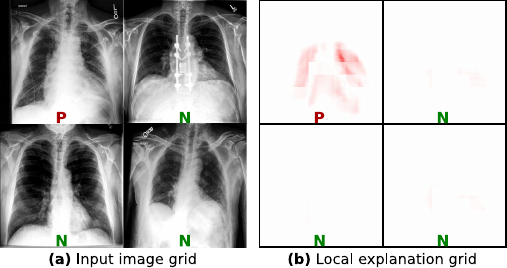}
  \caption{Example of image grid used for computing the class sensitivity metric for the disease cardiomegaly. \textbf{(a)} Input image grid with one positive and three negative samples. \textbf{(b)} Corresponding local explanations generated by Attri-Net.}
  \label{fig:img_grid}
\end{figure}

We evaluated the quality of the explanations using two metrics we defined: class sensitivity and disease sensitivity. 

We defined class sensitivity following the approach described by \cite{bohle2021convolutional}. Class sensitivity measures the intuition that explanations should be different for different image classes~\citep{khakzar2022explanations}. In our scenario, class sensitivity implies that the explanations for the disease-positive class should be distinct from those for the disease-negative class. Following \cite{bohle2021convolutional}, we generated a series of $2 \times 2$ grids of explanations, each containing only one positive example of a given disease (see example in Fig.~\ref{fig:img_grid}). The class sensitivity was then computed by dividing the sum of attributions in the positive example by the sum of all attributions in the grid. Ideally, all disease negative explanations should be blank because there is no disease effect in the negative samples, resulting in a class sensitivity score of $1$. Conversely, in the worst-case scenario where positive and negative sample explanations are indistinguishable, the class sensitivity score would be $\frac{1}{4}$. Similar to \cite{bohle2021convolutional}, we created 200 grids using the most confident positive and negative samples and computed the average class sensitivity across class $c$. In cases where certain diseases lacked an adequate number of positive samples in the test set, we constructed fewer image grids, with the number of grids equal to the count of correctly predicted positive examples.

We defined the disease sensitivity following the energy-based pointing game metric proposed by \cite{wang2020score}. Disease sensitivity measured how much of the attributions for a given disease are concentrated inside the ground truth bounding box or segmentation mask. It was computed by summing the attributions within the bounding box annotation and dividing by the sum of all attributions. The disease sensitivity scores were averaged across all classes $c$ and samples with ground truth pixel-wise annotations. Specifically, this included the remaining 60\% of bounding box annotated samples for the ChestX-ray8 dataset, as well as all samples in the test set for the CheXpert and VinDr-CXR datasets.

\subsection{Evaluation of Classification Performance}\label{sec:exp_classification}

\begin{table}[t]
    \centering
    \caption{Classification performance measured by area under the ROC curve (AUC).}
    \label{tab:auc} 
    \resizebox{\columnwidth}{!}{
    \begin{tabular}{p{4cm}|>{\centering\arraybackslash}p{1.5cm}>{\centering\arraybackslash}p{1.5cm}>{\centering\arraybackslash}p{1.5cm}}
        \hline
        \multirow{2}{*}{\bfseries Model}  & \bfseries CheXpert & \bfseries ChestX-ray8 &\bfseries VinDr-CXR\\
        \hline
        Stanford baseline
        \citep{irvin2019chexpert}
        & 0.907 & - & - \\
        DeepAUC \citep{yuan2021large} & 0.930 & - & - \\
        LSE \citep{ye2020weakly} & - & 0.755 & - \\
        ChestNet \citep{ye2020weakly} & - & 0.790 & - \\
        ResNet50 & \textbf{0.875} & 0.778 & 0.764   \\
        B-cos ResNet50 & 0.866  & 0.757 & \textbf{0.836} \\ 
        B-cos ResNet50 (guided) & 0.839 & 0.754 & 0.828 \\
        ours & 0.873 & \textbf{0.779} & 0.789 \\
        ours (pseudo guidance) & 0.848  & 0.774  & 0.773 \\
        ours (full guidance) & - & - & 0.782\\
  \hline
  \end{tabular}
  }
  \vspace{-0.3cm}
\end{table}

The classification outcomes of our model, the black-box model ResNet50, the inherently interpretable model B-cos ResNet50, and the guided variants of our model and B-cos ResNet50 are presented in Table \ref{tab:auc}. The disease-wise classification performance is provided in Table \ref{tab:auc_each_disease} in Appendix \ref{sec:appendix_classification}.
Additionally, we report the top-performing result in the CheXpert competition \citep{yuan2021large}, the baseline outcome outlined in the dataset paper \citep{irvin2019chexpert}, and published result from \citep{ye2020weakly} on ChestX-ray8. Attri-Net overall performed comparable to the state-of-the-art, with an AUC that was similar to other methods on CheXpert, 
slightly lower on Vindr-CXR, and slightly better on ChestX-ray8. In comparison to our prior work \citep{sun2023inherently}, both the ResNet50 model and our own model exhibited a decline in classification performance when assessed on the Vindr-CXR official test set. This is primarily due to the increased difficulty of the test set. Furthermore, we observed a slight decrease in classification performance with the guided versions of both our model and B-cos ResNet50. This suggested that the unguided versions may exploit class-irrelevant features to achieve optimal classification performance, whereas these features were restricted in the guided models. Additionally, on the Vindr-CXR dataset, the model trained with full guidance outperformed the model trained with pseudo guidance by a small amount, indicating that more precise guidance leads to improved performance.

\subsection{Evaluation of the Local Explanations}\label{sec:exp_local_explanation}

We derived local explanation for Attri-Net using the weighted attribution maps as detailed in Sec.\ref{sec:obtaining_local_explanations} and compared our local explanations with those from B-cos ResNet50 and five post-hoc methods that explain the prediction for the black-box ResNet50 model.

\begin{table*}
    \centering
    \caption{Comparison of class sensitivity and disease sensitivity.}
    \label{tab:class_and_disease_sensitivity}
    \begin{tabular}{l|ccc|ccc}
        \hline
        \multirow{2}{*} {\bfseries Model} & \multicolumn{3}{c|}{\textbf{Class sensitivity}}
        &\multicolumn{3}{c}{\textbf{Disease Sensitivity}}\\
        \bfseries  & \bfseries CheXpert & \bfseries ChestX-ray8 &\bfseries VinDr-CXR & \bfseries CheXpert & \bfseries ChestX-ray8 &\bfseries VinDr-CXR\\
        \hline 
        GB & 0.303 & 0.263 & 0.267 & 0.175 & 0.176  & 0.047 \\
        GCam & 0.191  & 0.225 & 0.176 & 0.192 & 0.125 & 0.061\\
        LIME & 0.254  & 0.229 &0.252 & 0.103 & 0.122 & 0.031 \\ 
        SHAP & 0.351 & 0.434 & 0.306 & 0.219 & 0.278 & 0.067\\
        Gifsp. & 0.300  & 0.688  & 0.317 & 0.191 &0.178  &0.052 \\
        B-cos ResNet50 & 0.266 & 0.276 & 0.240 & 0.259  & 0.235  & 0.089\\
        B-cos ResNet50 (guided)& 0.271 & 0.322 & 0.240 & 0.279 & 0.247  & 0.075\\
        ours & \textbf{0.684} & 0.951 & 0.862 & 0.207 & 0.158 & 0.075\\
        ours (pseudo guidance)  & 0.622 &\textbf{0.965} & \textbf{0.920} & \textbf{0.400} &\textbf{0.327}  & 0.156\\
        ours (full guidance)  & - & - & {0.872} & - & - & \textbf{0.204}\\
        \hline
    \end{tabular}
\end{table*}

\begin{figure*}[h]
    \centering
    \includegraphics[width=1.95\columnwidth]{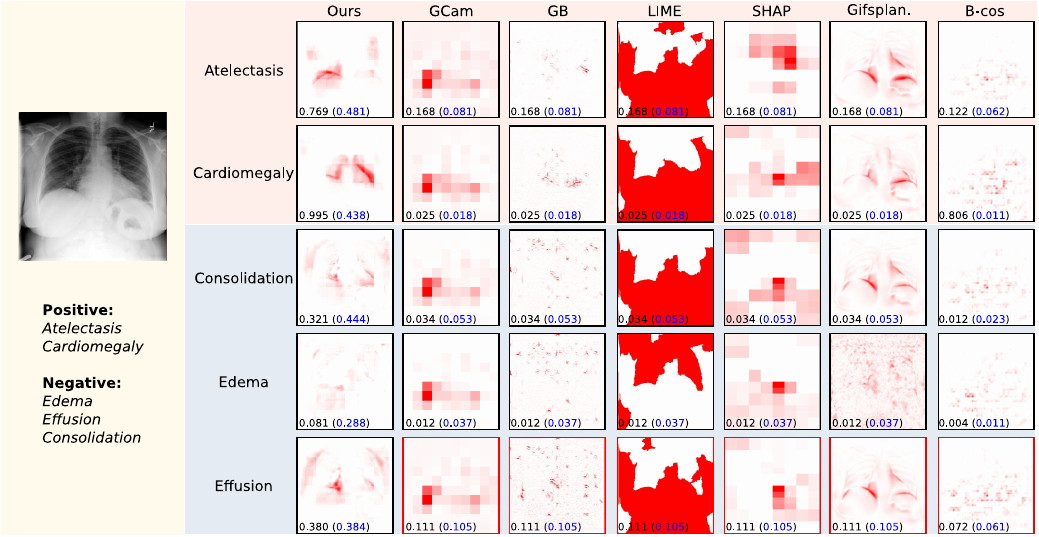}
    \caption{Visual comparison of explanations for an example image from the ChestX-ray8 dataset. Predicted class probabilities are indicated in the lower left corner of each attribution map with the respective decision threshold in parentheses. Wrong predictions are highlighted with red boxes.}
    \label{fig:all_exps}
\end{figure*}

\begin{figure}[!t]
\centerline{\includegraphics[width=1\columnwidth]{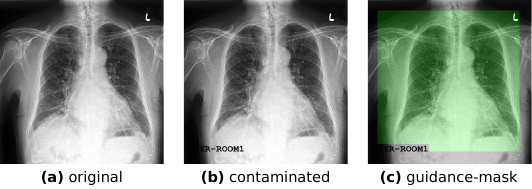}}
\vspace{-2mm}
\caption{Synthetically induced short-cut learning for global interpretability experiment. \textbf{(a)} A cardiomegaly positive sample from the CheXpert dataset. \textbf{(b)} The contaminated sample after adding the text ``CXR-ROOM1" as a spurious signal.  \textbf{(c)} The guidance mask encourages the model to use the features inside it and avoid using the spurious signal.}
\label{fig:guidance_shortcut}
\end{figure}

\subsubsection{Quantitative Analysis of Explanations}\label{sec:quantitative_results}

Table \ref{tab:class_and_disease_sensitivity} presents the class sensitivity and disease sensitivity of all local explanations that we compared. Additional disease-wise results along with the 95\% confidence interval are provided in Appendix \ref{sec:quantitative_results}.

Our method consistently outperformed all other methods, demonstrating significantly higher class sensitivity across all datasets. The high class sensitivity underscores Attri-Net's ability to provide more distinguishable explanations for disease-positive and disease-negative samples. 

Moreover, as depicted in Table \ref{tab:class_and_disease_sensitivity}, when trained with guidance, our Attri-Net achieved the highest disease sensitivity score across all datasets, indicating superior localization of disease-relevant regions compared to alternative methods. A paired t-test of our model trained with and without guidance on the CheXpert dataset reveals a statistically significant improvement when guidance was incorporated ($t = 28.284$, $p < 0.0001$). Additionally, the discrepancy in disease sensitivity between models trained with full guidance and those with pseudo guidance on VinDr-CXR emphasizes the importance of precise guidance in enhancing localization performance. Notably, similar to our Attri-Net, the disease sensitivity of the other inherently interpretable model, B-cos ResNet50, also improved after adding guidance on CheXpert and ChestX-ray8 datasets. However, the degree of improvement fell short of ours. This substantial performance gap highlighted that Attri-Net was particularly suitable for integrating guidance. This is because Attri-Net is explicitly designed for multi-label classification by distinguishing between positive and negative samples for each disease, resulting in more class-specific attribution maps. Therefore, the guidance effectively helps the model focuson the most relevant regions.

\subsubsection{Qualitative Analysis of Explanations}\label{sec:qualitative_results}

The qualitative examination of example explanations supported the quantitative results. Attri-Net was capable of generating local explanations that effectively emphasize the anatomical regions associated with the respective classes (see Fig.~\ref{fig:all_exps} for a representative example from the ChestX-ray8 dataset). Besides, explanations for highly confident predictions, such as cardiomegaly, exhibited a more pronounced disease effect compared to negative predictions such as Edema. Furthermore, the attributions for different classes were clearly distinct, each highlighting different anatomical areas. In contrast, it was challenging to understand the explanations from other post-hoc methods and the inherently interpretable baseline. For instance, explanations derived from Guided Backpropagation and B-cos ResNet50 were very noisy and hard to interpret. The explanations from the counterfactual-based Gifsplanation approach were easier to interpret, yet they consistently emphasized similar regions for different diseases. More examples from the CheXpert and VinDr-CXR datasets can be found in the Appendix \ref{sec:qualitative_results}. 

\begin{figure}
    \centering  
    \includegraphics[width=1\columnwidth]{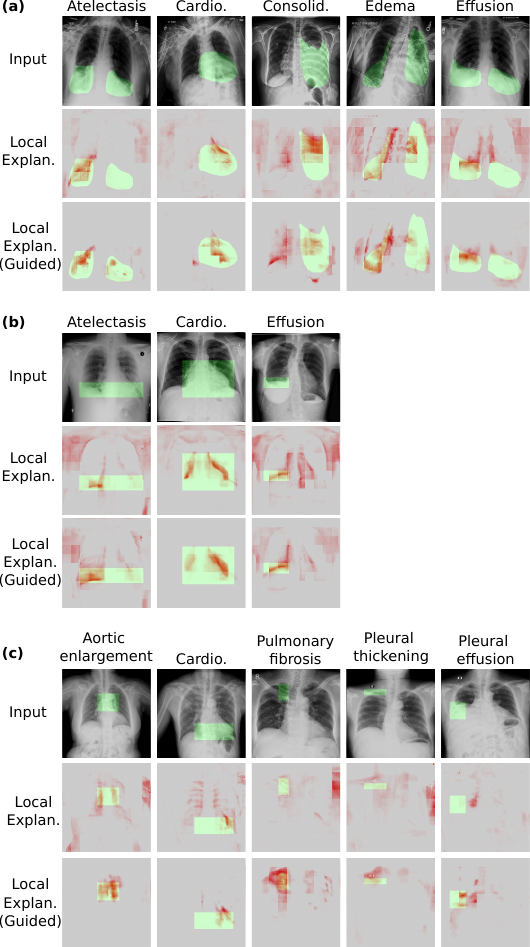}
    \vspace{-6mm}
    \caption{Local explanations produced by Attri-Net trained with and without guidance. Examples from three datasets: \textbf{(a)} CheXpert, \textbf{(b)} ChestX-ray8, \textbf{(c)} Vindr-CXR. Ground truth segmentation masks and bounding boxes are indicated in green regions.}
    \label{fig:all_exps_with_gt}
\end{figure}

We further examined Attri-Net explanations on images with pixel-wise ground truth annotations (Fig.~\ref{fig:all_exps_with_gt}) and observed that after adding pseudo-guidance to the model, the local explanations produced by Attri-Net better matched the areas where the pathologies are located, which was consistent with the high disease sensitivity in our quantitative evaluation. This suggested that incorporating guidance assisted the model in being ``right for the right reason."

\subsection{Evaluation of the Global Explanations}\label{sec:exp_global_explanation}

We assessed the performance of the global explanation mechanism of Attri-Net by investigating whether it can be used to uncover synthetically induced spurious correlation. 

Following our previously proposed evaluation framework \citep{sun2023right} we created a semi-synthetic dataset derived from the CheXpert dataset by contaminating 50\% of cardiomegaly positive samples with spurious tag signal (see Fig.~\ref{fig:guidance_shortcut} (a), (b)). This produces a spurious correlation between the tag and the presence of the disease cardiomegaly. 
In \citep{sun2023right} we showed that models trained on such a biased dataset will exhibit shortcut learning behavior, wherein they rely on spurious signals rather than relevant features for prediction. 

In this work, we trained the Attri-Net model on this contaminated dataset and qualitatively and quantitatively evaluated the model's reliance on the shortcut using the global explanation mechanism. Note that the other interpretable methods we assessed in Sec.\ref{sec:exp_local_explanation} lack the capability to offer global explanations. In fact Attri-Net is to our knowledge the first method to provide global spatial explanations. We thus limit our proof-of-concept to Attri-Net. 

In a second step, we investigated using model guidance to alleviate the model's shortcut learning behavior. Since we know the regions of spurious signals, we created guidance to avoid the spurious signal's region and encouraged the model to use the features inside the center region of the image (Fig.~\ref{fig:guidance_shortcut} (c)). We evaluated the global explanations for the model trained with guidance to verify whether the inclusion of guidance could reduce the model's dependence on spurious signals.

\subsubsection{Quantitative Results}\label{sec:global_quantitative}

Since we have introduce a known shortcut behaviour on the tag signal, we expect a global explanation to correctly highlight the reliance on the tag signal in contaminated test images. To measure this we have previously introduced the confounder sensitivity metric~\citep{sun2023right}. Confounder sensitivity measures how many of the top 10\% most significant pixels identified by the explanation fall on the synthetically added tag signal. We computed the confounder sensitivity using the positive class center of cardiomegaly, resulting in a score of 0.747. This indicates that 74.7\% of 
the spurious signal pixels were captured by the top 10\% most significant pixels in the positive class center of cardiomegaly.

To further verify that adding guidance can direct the model towards using the correct features, we added the guidance shown in Fig.~\ref{fig:guidance_shortcut}(c) to the model to encourage the model to avoid using spurious signals. After adding guidance, the confounder sensitivity of the new class center dropped to 0.002, which means that the most significant explanation pixels are not on the spurious signals anymore.

\subsubsection{Qualitative Results}\label{sec:global_qualitative}

\begin{figure}[h]
    \centering
    \includegraphics[width=1\columnwidth]{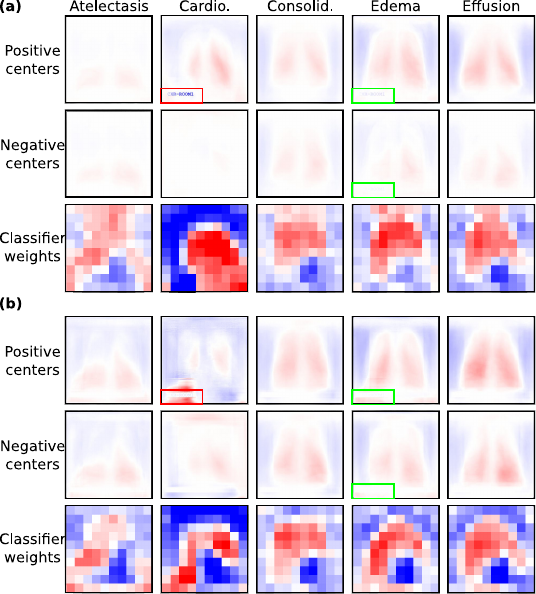}
    \caption{Global explanations produced by Attri-Net trained on the contaminated CheXpert dataset: (a) global explanation of the model trained without guidance, (b) global explanation of the model trained with guidance.  Red boxes highlight the spurious signal detected by the class centers of cardiomegaly, and green boxes highlight the spurious signal captured by the class centers of edema.}
    \label{fig:global_exp}
\end{figure}

We visualize the global explanation of the model trained on the contaminated CheXpert dataset in Fig.~\ref{fig:global_exp}(a) and assess if the spurious signal can be detected in the first place. From the positive class center of cardiomegaly, we can clearly see the text ``CXR-ROOM1'' which we added as the spurious signal. Meanwhile, we observed faintly discernible textual signals of edema within the class centers, indicating that Attri-Net used features from the spurious tag signal when diagnosing edema. Given that we only introduced the spurious signal to the positive class for cardiomegaly, this observation suggested a potential correlation between cardiomegaly and edema, agreeing with clinical observation in \citep{dodek1972pulmonary}. Since our model was trained to capture all information related to diagnosing disease, it was particularly useful for detecting faulty model behavior and the potential bias in the datasets. After adding model guidance, we found that the text was removed from the new global explanation (See Fig.~\ref{fig:global_exp}(b)), which indicated that the new model trained with guidance was less dependent on the spurious signal.

\subsection{Ablation study}\label{sec:ablation_study}

\begin{table*}
    \centering
    \caption{Ablation study on five losses. Evaluated on ChestX-ray8 dataset.}
    \label{tab:abla_loss} 
    \resizebox{\textwidth}{!}{
        \begin{tabular}{l|l|c|c|c}
            \hline
            \bfseries Model & \bfseries Loss terms & \bfseries Classification AUC 
            &\bfseries Class sensitivity &\bfseries Disease sensitivity\\
            \hline                  
            Attri-Net$_{\text{cls}}$ & $\mathcal{L}_{\text{cls}}$ 
            & 0.756 & 0.436 & 0.200\\
            Attri-Net$_{\text{cls\_adv}}$ & $\mathcal{L}_{\text{cls}}$ + $\mathcal{L}_{\text{adv}}$
            & 0.765 & 0.665 & 0.211\\
            Attri-Net$_{\text{cls\_adv\_reg}}$ & $\mathcal{L}_{\text{cls}}$ + $\mathcal{L}_{\text{adv}}$ + $\mathcal{L}_{\text{reg}}$
            & 0.778 & 0.954 & 0.178\\
           Attri-Net$_{\text{cls\_adv\_reg\_ctr}}$ & $\mathcal{L}_{\text{cls}}$ + $\mathcal{L}_{\text{adv}}$ + $\mathcal{L}_{\text{reg}}$+ $\mathcal{L}_{\text{ctr}}$
           & \textbf{0.779} & 0.951 & 0.158\\
           Attri-Net$_{\text{all}}$ & $\mathcal{L}_{\text{cls}}$ + $\mathcal{L}_{\text{adv}}$ + $\mathcal{L}_{\text{reg}}$ + $\mathcal{L}_{\text{ctr}}$ +
           $\mathcal{L}_{\text{guid}}$
           & 0.774 & \textbf{0.965} & \textbf{0.327}\\       
        \hline
    \end{tabular}
    }
  \vspace{-0.3cm}
\end{table*}

\begin{figure*}
    \centering
    \includegraphics[width=2\columnwidth]{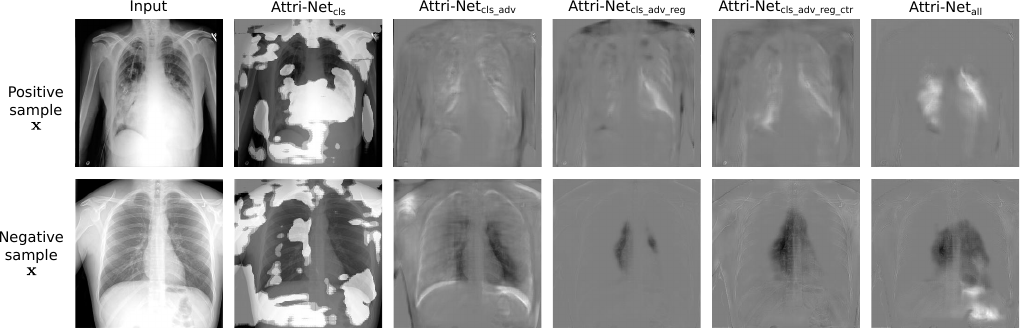}
    \caption{Qualitative results of ablation study.  Attribution maps generated by models trained with different subsets of our proposed losses for samples from the ChestX-ray8 dataset with cardiomegaly (top row) and without cardiomegaly (bottom row).}
    \label{fig:abla_loss}
\end{figure*}

We performed an ablation study on the five loss terms for training the class attribution generator network $M$ described in Sec.\ref{sec:methods}. In Fig.~\ref{fig:abla_loss}, we show examples of attribution maps from our model trained with different losses, and in Table \ref{tab:abla_loss}, we list the quantitative evaluation results of these models, respectively.

We found that models trained with different losses perform similarly in classification AUC. However, the attribution maps visually change greatly, and the quantitative evaluation results of class sensitivity and disease sensitivity vary a lot. As shown in Fig.~\ref{fig:abla_loss}, after adding the adversarial loss term $\mathcal{L}_{\text{adv}}$, the attribution maps focus on disease-relevant regions and become easy to understand. The regularization term $\mathcal{L}_{\text{reg}}$ encourages the attribution maps to be sparse. The addition of center loss $\mathcal{L}_{\text{ctr}}$ improves the classification slightly but does not substantially affect the attribution maps. More importantly, the class centers provide a possible way to interpret Attri-Net globally. From Table \ref{tab:abla_loss}, the model guidance loss $\mathcal{L}_{\text{guid}}$ greatly improves disease sensitivity, which is clearly shown from the attribution maps (last column) in Fig.~\ref{fig:abla_loss}.

\section{Discussion and Conclusion}\label{sec:discussion_and_conclusion}

We proposed Attri-Net, a novel inherently interpretable multi-label classifier that provides faithful local and global explanations. Our experiments showed that Attri-Net can produce high-quality local explanations that substantially outperform all baselines regarding class sensitivity and disease sensitivity while retaining classification performance comparable to state-of-the-art black-box models. 

Our further experiments showed that explanations of the black-box model can be highly dependent on which post-hoc technique is used, and fundamentally differ from each other even on the same sample. This erodes trust in their capacity to provide faithful explanations in high-stakes applications and shows the need for inherently interpretable models such as ours, where the predictions are formed directly and linearly from visually interpretable class attribution maps. 

Apart from providing transparent local explanations for individual predictions, Attri-Net can provide a global explanation of the model at the dataset level. Notably, the global explanation produced by Attri-Net was shown to faithfully capture synthetically induced shortcut learning behavior, which highlights the potential of our proposed approach to assist ML practitioners in improving the quality of the model and datasets. In addition, we demonstrated that Attri-Net can be combined with model guidance. If human annotations are available this allows to enforce that the correct features are used for making predictions and that the explanations are aligned with human knowledge. Our experiments showed that even very few human annotations can achieve significant improvement in disease localization.

The qualitative and quantitative assessments in this paper suggest that our method provides useful explanations; however, several limitations remain. Specifically, since Attri-Net performs classification and generates explanations by learning the differences between disease-positive and disease-negative classes, it is particularly suited for imaging modalities that capture relatively fixed anatomical structures and lesions. Therefore, detecting tiny tumors or nodules would be a challenging task for Attri-Net, especially when such lesions can appear in various locations. However, since many medical imaging modalities typically focus on specific parts of the human body with relatively fixed anatomical structures, the design concept of Attri-Net can potentially be extended to other imaging modalities, including 3D domains such as CT or MRI. In future work, we will evaluate the suitability of Attri-Net to other medical imaging domains with fixed structures (e.g. fundus imaging) as well as domains with varying structure (e.g. histopathological imaging). Since pixel-level lesion annotation often requires extensive expert effort, we will further explore the trade-off between annotation effort and the benefits of model guidance. Systematically quantifying global explanations remains a challenging task. In this study, the shortcut learning scenarios used for evaluation are limited to systematic artifacts with relatively fixed spatial locations. However, in real-world settings, shortcuts can arise in more diverse and spatially variable patterns. In future work, we aim to investigate more complex shortcut learning conditions and develop improved methods for quantifying global explanations.
Finally, we believe it is also crucial to assess the utility of Attri-Net and other explanation methods in human-in-the-loop settings, which is a crucial step toward clinical deployment.




%



\acks{This work was funded by the Deutsche Forschungsgemeinschaft (DFG) – EXC number 2064/1 – Project number 390727645 and the Carl Zeiss Foundation in the project ``Certification and Foundations of Safe Machine Learning Systems in Healthcare". The authors acknowledge the support of the Hertie Foundation. The authors thank the International Max Planck Research School for Intelligent Systems (IMPRS-IS) for supporting Susu Sun and Stefano Woerner.}

%
\ethics{The work follows appropriate ethical standards in conducting research and writing the manuscript, following all applicable laws and regulations regarding treatment of animals or human subjects.}

\coi{We declare we don't have conflicts of interest.}


\data{We used public datasets for our experiments, and we made the code for reproducing the results publicly available at \url{ https://github.com/ss-sun/Attri-Net-V2}.}

\bibliography{sample}

@article{litjens2017survey,
  title={A survey on deep learning in medical image analysis},
  author={Litjens, Geert and Kooi, Thijs and Bejnordi, Babak Ehteshami and Setio, Arnaud Arindra Adiyoso and Ciompi, Francesco and Ghafoorian, Mohsen and Van Der Laak, Jeroen Awm and Van Ginneken, Bram and S{\'a}nchez, Clara I},
  journal={Medical image analysis},
  volume={42},
  pages={60--88},
  year={2017},
  publisher={Elsevier}
}

@article{dietvorst2015algorithm,
  title={Algorithm aversion: people erroneously avoid algorithms after seeing them err.},
  author={Dietvorst, Berkeley J and Simmons, Joseph P and Massey, Cade},
  journal={Journal of Experimental Psychology: General},
  volume={144},
  number={1},
  pages={114},
  year={2015},
  publisher={American Psychological Association}
}

@article{tschandl2020human,
  title={Human--computer collaboration for skin cancer recognition},
  author={Tschandl, Philipp and Rinner, Christoph and Apalla, Zoe and Argenziano, Giuseppe and Codella, Noel and Halpern, Allan and Janda, Monika and Lallas, Aimilios and Longo, Caterina and Malvehy, Josep and others},
  journal={Nature Medicine},
  volume={26},
  number={8},
  pages={1229--1234},
  year={2020},
  publisher={Nature Publishing Group US New York}
}

@article{gaube2021ai,
  title={Do as AI say: susceptibility in deployment of clinical decision-aids},
  author={Gaube, Susanne and Suresh, Harini and Raue, Martina and Merritt, Alexander and Berkowitz, Seth J and Lermer, Eva and Coughlin, Joseph F and Guttag, John V and Colak, Errol and Ghassemi, Marzyeh},
  journal={NPJ digital medicine},
  volume={4},
  number={1},
  pages={31},
  year={2021},
  publisher={Nature Publishing Group UK London}
}

@article{grote2020ethics,
  title={On the ethics of algorithmic decision-making in healthcare},
  author={Grote, Thomas and Berens, Philipp},
  journal={Journal of medical ethics},
  volume={46},
  number={3},
  pages={205--211},
  year={2020},
  publisher={Institute of Medical Ethics}
}

@article{rudin2019stop,
  title={Stop explaining black box machine learning models for high stakes decisions and use interpretable models instead},
  author={Rudin, Cynthia},
  journal={Nature machine intelligence},
  volume={1},
  number={5},
  pages={206--215},
  year={2019},
  publisher={Nature Publishing Group UK London}
}

@article{adebayo2018sanity,
  title={Sanity checks for saliency maps},
  author={Adebayo, Julius and Gilmer, Justin and Muelly, Michael and Goodfellow, Ian and Hardt, Moritz and Kim, Been},
  journal={Advances in neural information processing systems},
  volume={31},
  year={2018}
}

@article{arun2021assessing,
  title={Assessing the trustworthiness of saliency maps for localizing abnormalities in medical imaging},
  author={Arun, Nishanth and Gaw, Nathan and Singh, Praveer and Chang, Ken and Aggarwal, Mehak and Chen, Bryan and Hoebel, Katharina and Gupta, Sharut and Patel, Jay and Gidwani, Mishka and others},
  journal={Radiology: Artificial Intelligence},
  volume={3},
  number={6},
  pages={e200267},
  year={2021},
  publisher={Radiological Society of North America}
}

@article{chen2019looks,
  title={This looks like that: deep learning for interpretable image recognition},
  author={Chen, Chaofan and Li, Oscar and Tao, Daniel and Barnett, Alina and Rudin, Cynthia and Su, Jonathan K},
  journal={Advances in neural information processing systems},
  volume={32},
  year={2019}
}

@misc{pihlgren2025segmentationsmoothingaffectexplanation,
      title={Segmentation and Smoothing Affect Explanation Quality More Than the Choice of Perturbation-based XAI Method for Image Explanations}, 
      author={Gustav Grund Pihlgren and Kary Främling},
      year={2025},
      eprint={2409.04116},
      archivePrefix={arXiv},
      primaryClass={cs.CV},
      url={https://arxiv.org/abs/2409.04116}, 
}

@inproceedings{koh2020concept,
  title={Concept bottleneck models},
  author={Koh, Pang Wei and Nguyen, Thao and Tang, Yew Siang and Mussmann, Stephen and Pierson, Emma and Kim, Been and Liang, Percy},
  booktitle={International conference on machine learning},
  pages={5338--5348},
  year={2020},
  organization={PMLR}
}

@article{bohle2024b,
  title={B-cos Alignment for Inherently Interpretable CNNs and Vision Transformers},
  author={B{\"o}hle, Moritz and Singh, Navdeeppal and Fritz, Mario and Schiele, Bernt},
  journal={IEEE Transactions on Pattern Analysis and Machine Intelligence},
  year={2024},
  publisher={IEEE}
}

@article{geirhos2020shortcut,
  title={Shortcut learning in deep neural networks},
  author={Geirhos, Robert and Jacobsen, J{\"o}rn-Henrik and Michaelis, Claudio and Zemel, Richard and Brendel, Wieland and Bethge, Matthias and Wichmann, Felix A},
  journal={Nature Machine Intelligence},
  volume={2},
  number={11},
  pages={665--673},
  year={2020},
  publisher={Nature Publishing Group UK London}
}

@article{sagawa2019distributionally,
  title={Distributionally robust neural networks for group shifts: On the importance of regularization for worst-case generalization},
  author={Sagawa, Shiori and Koh, Pang Wei and Hashimoto, Tatsunori B and Liang, Percy},
  journal={arXiv preprint arXiv:1911.08731},
  year={2019}
}

@article{erion2021improving,
  title={Improving performance of deep learning models with axiomatic attribution priors and expected gradients},
  author={Erion, Gabriel and Janizek, Joseph D and Sturmfels, Pascal and Lundberg, Scott M and Lee, Su-In},
  journal={Nature machine intelligence},
  volume={3},
  number={7},
  pages={620--631},
  year={2021},
  publisher={Nature Publishing Group UK London}
}

@inproceedings{pillai2022consistent,
  title={Consistent explanations by contrastive learning},
  author={Pillai, Vipin and Koohpayegani, Soroush Abbasi and Ouligian, Ashley and Fong, Dennis and Pirsiavash, Hamed},
  booktitle={Proceedings of the IEEE/CVF Conference on Computer Vision and Pattern Recognition},
  pages={10213--10222},
  year={2022}
}

@inproceedings{zhou2016learning,
  title={Learning deep features for discriminative localization},
  author={Zhou, Bolei and Khosla, Aditya and Lapedriza, Agata and Oliva, Aude and Torralba, Antonio},
  booktitle={Proceedings of the IEEE conference on computer vision and pattern recognition},
  pages={2921--2929},
  year={2016}
}

@article{brendel2019approximating,
  title={Approximating cnns with bag-of-local-features models works surprisingly well on imagenet},
  author={Brendel, Wieland and Bethge, Matthias},
  journal={arXiv preprint arXiv:1904.00760},
  year={2019}
}

@inproceedings{donteu2023sparse,
  title={Sparse Activations for Interpretable Disease Grading},
  author={Donteu, Kerol R Djoumessi and Ilanchezian, Indu and K{\"u}hlewein, Laura and Faber, Hanna and Baumgartner, Christian F and Bah, Bubacarr and Berens, Philipp and Koch, Lisa M},
  booktitle={Medical Imaging with Deep Learning},
  year={2023}
}

@article{schutte2021using,
  title={Using stylegan for visual interpretability of deep learning models on medical images},
  author={Schutte, Kathryn and Moindrot, Olivier and H{\'e}rent, Paul and Schiratti, Jean-Baptiste and J{\'e}gou, Simon},
  journal={arXiv preprint arXiv:2101.07563},
  year={2021}
}

@article{joshi2018xgems,
  title={xgems: Generating examplars to explain black-box models},
  author={Joshi, Shalmali and Koyejo, Oluwasanmi and Kim, Been and Ghosh, Joydeep},
  journal={arXiv preprint arXiv:1806.08867},
  year={2018}
}

@article{singla2019explanation,
  title={Explanation by progressive exaggeration},
  author={Singla, Sumedha and Pollack, Brian and Chen, Junxiang and Batmanghelich, Kayhan},
  journal={arXiv preprint arXiv:1911.00483},
  year={2019}
}

@inproceedings{pinheiro2015image,
  title={From image-level to pixel-level labeling with convolutional networks},
  author={Pinheiro, Pedro O and Collobert, Ronan},
  booktitle={Proceedings of the IEEE conference on computer vision and pattern recognition},
  pages={1713--1721},
  year={2015}
}

@inproceedings{zhu2017deep,
  title={Deep multi-instance networks with sparse label assignment for whole mammogram classification},
  author={Zhu, Wentao and Lou, Qi and Vang, Yeeleng Scott and Xie, Xiaohui},
  booktitle={Medical Image Computing and Computer Assisted Intervention- MICCAI 2017: 20th International Conference, Quebec City, QC, Canada, September 11-13, 2017, Proceedings, Part III 20},
  pages={603--611},
  year={2017},
  organization={Springer}
}

@article{alvarez2018towards,
  title={Towards robust interpretability with self-explaining neural networks},
  author={Alvarez Melis, David and Jaakkola, Tommi},
  journal={Advances in neural information processing systems},
  volume={31},
  year={2018}
}

@article{chen2020concept,
  title={Concept whitening for interpretable image recognition},
  author={Chen, Zhi and Bei, Yijie and Rudin, Cynthia},
  journal={Nature Machine Intelligence},
  volume={2},
  number={12},
  pages={772--782},
  year={2020},
  publisher={Nature Publishing Group UK London}
}

@article{barnett2021interpretable,
  title={Interpretable mammographic image classification using case-based reasoning and deep learning},
  author={Barnett, Alina Jade and Schwartz, Fides Regina and Tao, Chaofan and Chen, Chaofan and Ren, Yinhao and Lo, Joseph Y and Rudin, Cynthia},
  journal={arXiv preprint arXiv:2107.05605},
  year={2021}
}

@inproceedings{arjovsky2017wasserstein,
  title={Wasserstein generative adversarial networks},
  author={Arjovsky, Martin and Chintala, Soumith and Bottou, L{\'e}on},
  booktitle={International conference on machine learning},
  pages={214--223},
  year={2017},
  organization={PMLR}
}

@inproceedings{sun2021task,
  title={Task switching network for multi-task learning},
  author={Sun, Guolei and Probst, Thomas and Paudel, Danda Pani and Popovi{\'c}, Nikola and Kanakis, Menelaos and Patel, Jagruti and Dai, Dengxin and Van Gool, Luc},
  booktitle={Proceedings of the IEEE/CVF international conference on computer vision},
  pages={8291--8300},
  year={2021}
}

@inproceedings{wen2016discriminative,
  title={A discriminative feature learning approach for deep face recognition},
  author={Wen, Yandong and Zhang, Kaipeng and Li, Zhifeng and Qiao, Yu},
  booktitle={Computer vision--ECCV 2016: 14th European conference, amsterdam, the netherlands, October 11--14, 2016, proceedings, part VII 14},
  pages={499--515},
  year={2016},
  organization={Springer}
}

@inproceedings{choi2018stargan,
  title={Stargan: Unified generative adversarial networks for multi-domain image-to-image translation},
  author={Choi, Yunjey and Choi, Minje and Kim, Munyoung and Ha, Jung-Woo and Kim, Sunghun and Choo, Jaegul},
  booktitle={Proceedings of the IEEE conference on computer vision and pattern recognition},
  pages={8789--8797},
  year={2018}
}

@article{kingma2014adam,
  title={Adam: A method for stochastic optimization},
  author={Kingma, Diederik P and Ba, Jimmy},
  journal={arXiv preprint arXiv:1412.6980},
  year={2014}
}

@inproceedings{lakkaraju2016interpretable,
  title={Interpretable decision sets: A joint framework for description and prediction},
  author={Lakkaraju, Himabindu and Bach, Stephen H and Leskovec, Jure},
  booktitle={Proceedings of the 22nd ACM SIGKDD international conference on knowledge discovery and data mining},
  pages={1675--1684},
  year={2016}
}

@article{angelino2018learning,
  title={Learning certifiably optimal rule lists for categorical data},
  author={Angelino, Elaine and Larus-Stone, Nicholas and Alabi, Daniel and Seltzer, Margo and Rudin, Cynthia},
  journal={Journal of Machine Learning Research},
  volume={18},
  number={234},
  pages={1--78},
  year={2018}
}

@article{baumgartner2017sononet,
  title={SonoNet: real-time detection and localisation of fetal standard scan planes in freehand ultrasound},
  author={Baumgartner, Christian F and Kamnitsas, Konstantinos and Matthew, Jacqueline and Fletcher, Tara P and Smith, Sandra and Koch, Lisa M and Kainz, Bernhard and Rueckert, Daniel},
  journal={IEEE transactions on medical imaging},
  volume={36},
  number={11},
  pages={2204--2215},
  year={2017},
  publisher={IEEE}
}

@inproceedings{jamaludin2016spinenet,
  title={SpineNet: automatically pinpointing classification evidence in spinal MRIs},
  author={Jamaludin, Amir and Kadir, Timor and Zisserman, Andrew},
  booktitle={International Conference on Medical Image Computing and Computer-Assisted Intervention},
  pages={166--175},
  year={2016},
  organization={Springer}
}

@article{springenberg2014striving,
  title={Striving for simplicity: The all convolutional net},
  author={Springenberg, Jost Tobias and Dosovitskiy, Alexey and Brox, Thomas and Riedmiller, Martin},
  journal={arXiv preprint arXiv:1412.6806},
  year={2014}
}

@inproceedings{ribeiro2016should,
  title={" Why should i trust you?" Explaining the predictions of any classifier},
  author={Ribeiro, Marco Tulio and Singh, Sameer and Guestrin, Carlos},
  booktitle={Proceedings of the 22nd ACM SIGKDD international conference on knowledge discovery and data mining},
  pages={1135--1144},
  year={2016}
}

@article{lundberg2017unified,
  title={A unified approach to interpreting model predictions},
  author={Lundberg, Scott M and Lee, Su-In},
  journal={Advances in neural information processing systems},
  volume={30},
  year={2017}
}

@article{bank2023autoencoders,
  title={Autoencoders},
  author={Bank, Dor and Koenigstein, Noam and Giryes, Raja},
  journal={Machine learning for data science handbook: data mining and knowledge discovery handbook},
  pages={353--374},
  year={2023},
  publisher={Springer}
}

@inproceedings{cohen2021gifsplanation,
  title={Gifsplanation via latent shift: a simple autoencoder approach to counterfactual generation for chest x-rays},
  author={Cohen, Joseph Paul and Brooks, Rupert and En, Sovann and Zucker, Evan and Pareek, Anuj and Lungren, Matthew P and Chaudhari, Akshay},
  booktitle={Medical Imaging with Deep Learning},
  pages={74--104},
  year={2021},
  organization={PMLR}
}

@misc{cohen2025identifyingspuriouscorrelationsusing,
      title={Identifying Spurious Correlations using Counterfactual Alignment}, 
      author={Joseph Paul Cohen and Louis Blankemeier and Akshay Chaudhari},
      year={2025},
      eprint={2312.02186},
      archivePrefix={arXiv},
      primaryClass={cs.CV},
      url={https://arxiv.org/abs/2312.02186}, 
}

@article{ho2020denoising,
  title={Denoising diffusion probabilistic models},
  author={Ho, Jonathan and Jain, Ajay and Abbeel, Pieter},
  journal={Advances in neural information processing systems},
  volume={33},
  pages={6840--6851},
  year={2020}
}

@article{kim2016examples,
  title={Examples are not enough, learn to criticize! criticism for interpretability},
  author={Kim, Been and Khanna, Rajiv and Koyejo, Oluwasanmi O},
  journal={Advances in neural information processing systems},
  volume={29},
  year={2016}
}

@article{ghassemi2021false,
  title={The false hope of current approaches to explainable artificial intelligence in health care},
  author={Ghassemi, Marzyeh and Oakden-Rayner, Luke and Beam, Andrew L},
  journal={The Lancet Digital Health},
  volume={3},
  number={11},
  pages={e745--e750},
  year={2021},
  publisher={Elsevier}
}

@inproceedings{baumgartner2018visual,
  title={Visual feature attribution using wasserstein gans},
  author={Baumgartner, Christian F and Koch, Lisa M and Tezcan, Kerem Can and Ang, Jia Xi and Konukoglu, Ender},
  booktitle={Proceedings of the IEEE conference on computer vision and pattern recognition},
  pages={8309--8319},
  year={2018}
}

@article{sun2023inherently,
  title={Inherently interpretable multi-label classification using class-specific counterfactuals},
  author={Sun, Susu and Woerner, Stefano and Maier, Andreas and Koch, Lisa M and Baumgartner, Christian F},
  journal={arXiv preprint arXiv:2303.00500},
  year={2023}
}

@inproceedings{sun2023right,
  title={Right for the Wrong Reason: Can Interpretable ML Techniques Detect Spurious Correlations?},
  author={Sun, Susu and Koch, Lisa M and Baumgartner, Christian F},
  booktitle={International Conference on Medical Image Computing and Computer-Assisted Intervention},
  pages={425--434},
  year={2023},
  organization={Springer}
}

@inproceedings{irvin2019chexpert,
  title={Chexpert: A large chest radiograph dataset with uncertainty labels and expert comparison},
  author={Irvin, Jeremy and Rajpurkar, Pranav and Ko, Michael and Yu, Yifan and Ciurea-Ilcus, Silviana and Chute, Chris and Marklund, Henrik and Haghgoo, Behzad and Ball, Robyn and Shpanskaya, Katie and others},
  booktitle={Proceedings of the AAAI conference on artificial intelligence},
  volume={33},
  number={01},
  pages={590--597},
  year={2019}
}

@inproceedings{yuan2021large,
  title={Large-scale robust deep auc maximization: A new surrogate loss and empirical studies on medical image classification},
  author={Yuan, Zhuoning and Yan, Yan and Sonka, Milan and Yang, Tianbao},
  booktitle={Proceedings of the IEEE/CVF International Conference on Computer Vision},
  pages={3040--3049},
  year={2021}
}

@article{ye2020weakly,
  title={Weakly supervised lesion localization with probabilistic-cam pooling},
  author={Ye, Wenwu and Yao, Jin and Xue, Hui and Li, Yi},
  journal={arXiv preprint arXiv:2005.14480},
  year={2020}
}

@inproceedings{wang2017chestx,
  title={Chestx-ray8: Hospital-scale chest x-ray database and benchmarks on weakly-supervised classification and localization of common thorax diseases},
  author={Wang, Xiaosong and Peng, Yifan and Lu, Le and Lu, Zhiyong and Bagheri, Mohammadhadi and Summers, Ronald M},
  booktitle={Proceedings of the IEEE conference on computer vision and pattern recognition},
  pages={2097--2106},
  year={2017}
}

@article{nguyen2022vindr,
  title={VinDr-CXR: An open dataset of chest X-rays with radiologist’s annotations},
  author={Nguyen, Ha Q and Lam, Khanh and Le, Linh T and Pham, Hieu H and Tran, Dat Q and Nguyen, Dung B and Le, Dung D and Pham, Chi M and Tong, Hang TT and Dinh, Diep H and others},
  journal={Scientific Data},
  volume={9},
  number={1},
  pages={429},
  year={2022},
  publisher={Nature Publishing Group UK London}
}

@article{saporta2022benchmarking,
  title={Benchmarking saliency methods for chest X-ray interpretation},
  author={Saporta, Adriel and Gui, Xiaotong and Agrawal, Ashwin and Pareek, Anuj and Truong, Steven QH and Nguyen, Chanh DT and Ngo, Van-Doan and Seekins, Jayne and Blankenberg, Francis G and Ng, Andrew Y and others},
  journal={Nature Machine Intelligence},
  volume={4},
  number={10},
  pages={867--878},
  year={2022},
  publisher={Nature Publishing Group UK London}
}

@inproceedings{li2018thoracic,
  title={Thoracic disease identification and localization with limited supervision},
  author={Li, Zhe and Wang, Chong and Han, Mei and Xue, Yuan and Wei, Wei and Li, Li-Jia and Fei-Fei, Li},
  booktitle={Proceedings of the IEEE conference on computer vision and pattern recognition},
  pages={8290--8299},
  year={2018}
}

@inproceedings{he2016deep,
  title={Deep residual learning for image recognition},
  author={He, Kaiming and Zhang, Xiangyu and Ren, Shaoqing and Sun, Jian},
  booktitle={Proceedings of the IEEE conference on computer vision and pattern recognition},
  pages={770--778},
  year={2016}
}

@inproceedings{selvaraju2017grad,
  title={Grad-cam: Visual explanations from deep networks via gradient-based localization},
  author={Selvaraju, Ramprasaath R and Cogswell, Michael and Das, Abhishek and Vedantam, Ramakrishna and Parikh, Devi and Batra, Dhruv},
  booktitle={Proceedings of the IEEE international conference on computer vision},
  pages={618--626},
  year={2017}
}

@inproceedings{bohle2021convolutional,
  title={Convolutional dynamic alignment networks for interpretable classifications},
  author={Bohle, Moritz and Fritz, Mario and Schiele, Bernt},
  booktitle={Proceedings of the IEEE/CVF Conference on Computer Vision and Pattern Recognition},
  pages={10029--10038},
  year={2021}
}

@inproceedings{wang2020score,
  title={Score-CAM: Score-weighted visual explanations for convolutional neural networks},
  author={Wang, Haofan and Wang, Zifan and Du, Mengnan and Yang, Fan and Zhang, Zijian and Ding, Sirui and Mardziel, Piotr and Hu, Xia},
  booktitle={Proceedings of the IEEE/CVF conference on computer vision and pattern recognition workshops},
  pages={24--25},
  year={2020}
}

@inproceedings{shrikumar2017learning,
  title={Learning important features through propagating activation differences},
  author={Shrikumar, Avanti and Greenside, Peyton and Kundaje, Anshul},
  booktitle={International conference on machine learning},
  pages={3145--3153},
  year={2017},
  organization={PMLR}
}

@inproceedings{sundararajan2017axiomatic,
  title={Axiomatic attribution for deep networks},
  author={Sundararajan, Mukund and Taly, Ankur and Yan, Qiqi},
  booktitle={International conference on machine learning},
  pages={3319--3328},
  year={2017},
  organization={PMLR}
}

@article{reyes2020interpretability,
  title={On the interpretability of artificial intelligence in radiology: challenges and opportunities},
  author={Reyes, Mauricio and Meier, Raphael and Pereira, S{\'e}rgio and Silva, Carlos A and Dahlweid, Fried-Michael and Tengg-Kobligk, Hendrik von and Summers, Ronald M and Wiest, Roland},
  journal={Radiology: artificial intelligence},
  volume={2},
  number={3},
  pages={e190043},
  year={2020},
  publisher={Radiological Society of North America}
}

@inproceedings{kim2018interpretability,
  title={Interpretability beyond feature attribution: Quantitative testing with concept activation vectors (tcav)},
  author={Kim, Been and Wattenberg, Martin and Gilmer, Justin and Cai, Carrie and Wexler, James and Viegas, Fernanda and others},
  booktitle={International conference on machine learning},
  pages={2668--2677},
  year={2018},
  organization={PMLR}
}

@article{pereira2018enhancing,
  title={Enhancing interpretability of automatically extracted machine learning features: application to a RBM-Random Forest system on brain lesion segmentation},
  author={Pereira, S{\'e}rgio and Meier, Raphael and McKinley, Richard and Wiest, Roland and Alves, Victor and Silva, Carlos A and Reyes, Mauricio},
  journal={Medical image analysis},
  volume={44},
  pages={228--244},
  year={2018},
  publisher={Elsevier}
}

@inproceedings{khakzar2022explanations,
  title={Do explanations explain? Model knows best},
  author={Khakzar, Ashkan and Khorsandi, Pedram and Nobahari, Rozhin and Navab, Nassir},
  booktitle={Proceedings of the IEEE/CVF Conference on Computer Vision and Pattern Recognition},
  pages={10244--10253},
  year={2022}
}

@article{dodek1972pulmonary,
  title={Pulmonary edema in coronary-artery disease without cardiomegaly: paradox of the stiff heart},
  author={Dodek, Arthur and Kassebaum, Donald G and Bristow, J David},
  journal={New England Journal of Medicine},
  volume={286},
  number={25},
  pages={1347--1350},
  year={1972},
  publisher={Mass Medical Soc}
}

@article{zhang2019should,
  title={"Why should you trust my explanation?" understanding uncertainty in LIME explanations},
  author={Zhang, Yujia and Song, Kuangyan and Sun, Yiming and Tan, Sarah and Udell, Madeleine},
  journal={arXiv preprint arXiv:1904.12991},
  year={2019}
}

@article{chen2018shapley,
  title={L-shapley and c-shapley: Efficient model interpretation for structured data},
  author={Chen, Jianbo and Song, Le and Wainwright, Martin J and Jordan, Michael I},
  journal={arXiv preprint arXiv:1808.02610},
  year={2018}
}

@article{draelos2020use,
  title={Use HiResCAM instead of Grad-CAM for faithful explanations of convolutional neural networks},
  author={Draelos, Rachel Lea and Carin, Lawrence},
  journal={arXiv preprint arXiv:2011.08891},
  year={2020}
}

@inproceedings{boreiko2022visual,
  title={Visual explanations for the detection of diabetic retinopathy from retinal fundus images},
  author={Boreiko, Valentyn and Ilanchezian, Indu and Ayhan, Murat Se{\c{c}}kin and M{\"u}ller, Sarah and Koch, Lisa M and Faber, Hanna and Berens, Philipp and Hein, Matthias},
  booktitle={International conference on medical image computing and computer-assisted intervention},
  pages={539--549},
  year={2022},
  organization={Springer}
}

@article{snell2017prototypical,
  title={Prototypical networks for few-shot learning},
  author={Snell, Jake and Swersky, Kevin and Zemel, Richard},
  journal={Advances in neural information processing systems},
  volume={30},
  year={2017}
}

@article{fathi2024decodex,
  title={DeCoDEx: Confounder Detector Guidance for Improved Diffusion-based Counterfactual Explanations},
  author={Fathi, Nima and Kumar, Amar and Nichyporuk, Brennan and Havaei, Mohammad and Arbel, Tal},
  journal={arXiv preprint arXiv:2405.09288},
  year={2024}
}

@article{goodfellow2020generative,
  title={Generative adversarial networks},
  author={Goodfellow, Ian and Pouget-Abadie, Jean and Mirza, Mehdi and Xu, Bing and Warde-Farley, David and Ozair, Sherjil and Courville, Aaron and Bengio, Yoshua},
  journal={Communications of the ACM},
  volume={63},
  number={11},
  pages={139--144},
  year={2020},
  publisher={ACM New York, NY, USA}
}

@article{atad2022chexplaining,
  title={Chexplaining in style: Counterfactual explanations for chest x-rays using stylegan},
  author={Atad, Matan and Dmytrenko, Vitalii and Li, Yitong and Zhang, Xinyue and Keicher, Matthias and Kirschke, Jan and Wiestler, Bene and Khakzar, Ashkan and Navab, Nassir},
  journal={arXiv preprint arXiv:2207.07553},
  year={2022}
}

@article{qi2025projectedex,
  title={Projectedex: Enhancing generation in explainable ai for prostate cancer},
  author={Qi, Xuyin and Zhang, Zeyu and Handoko, Aaron Berliano and Zheng, Huazhan and Chen, Mingxi and Huy, Ta Duc and Phan, Vu Minh Hieu and Zhang, Lei and Cheng, Linqi and Jiang, Shiyu and others},
  journal={arXiv preprint arXiv:2501.01392},
  year={2025}
}

@inproceedings{mindlin2023abc,
  title={ABC-GAN: Spatially Constrained Counterfactual Generation for Image Classification Explanations},
  author={Mindlin, Dimitry and Schilling, Malte and Cimiano, Philipp},
  booktitle={World Conference on Explainable Artificial Intelligence},
  pages={260--282},
  year={2023},
  organization={Springer}
}

@article{garg2024advancing,
  title={Advancing ante-hoc explainable models through generative adversarial networks},
  author={Garg, Tanmay and Vemuri, Deepika and Balasubramanian, Vineeth N},
  journal={arXiv preprint arXiv:2401.04647},
  year={2024}
}

@inproceedings{jeanneret2023adversarial,
  title={Adversarial counterfactual visual explanations},
  author={Jeanneret, Guillaume and Simon, Lo{\"\i}c and Jurie, Fr{\'e}d{\'e}ric},
  booktitle={Proceedings of the IEEE/CVF Conference on Computer Vision and Pattern Recognition},
  pages={16425--16435},
  year={2023}
}

@article{bedel2024dreamr,
  title={Dreamr: Diffusion-driven counterfactual explanation for functional mri},
  author={Bedel, Hasan A and {\c{C}}ukur, Tolga},
  journal={IEEE Transactions on Medical Imaging},
  year={2024},
  publisher={IEEE}
}

@article{bassan2024local,
  title={Local vs. Global Interpretability: A Computational Complexity Perspective},
  author={Bassan, Shahaf and Amir, Guy and Katz, Guy},
  journal={arXiv preprint arXiv:2406.02981},
  year={2024}
}

@article{christoph2020interpretable,
  title={Interpretable machine learning: A guide for making black box models explainable},
  author={Christoph, Molnar},
  year={2020},
  publisher={Leanpub}
}

@article{augustin2022diffusion,
  title={Diffusion visual counterfactual explanations},
  author={Augustin, Maximilian and Boreiko, Valentyn and Croce, Francesco and Hein, Matthias},
  journal={Advances in Neural Information Processing Systems},
  volume={35},
  pages={364--377},
  year={2022}
}

@inproceedings{rao2023studying,
  title={Studying how to efficiently and effectively guide models with explanations},
  author={Rao, Sukrut and B{\"o}hle, Moritz and Parchami-Araghi, Amin and Schiele, Bernt},
  booktitle={Proceedings of the IEEE/CVF International Conference on Computer Vision},
  pages={1922--1933},
  year={2023}
}


\clearpage
\appendix




\section{Additional examples of counterfactual attribution maps}\label{sec:examples_counterfactual}

\begin{figure*}[h]
    \centering
    \includegraphics[width=1.8\columnwidth]{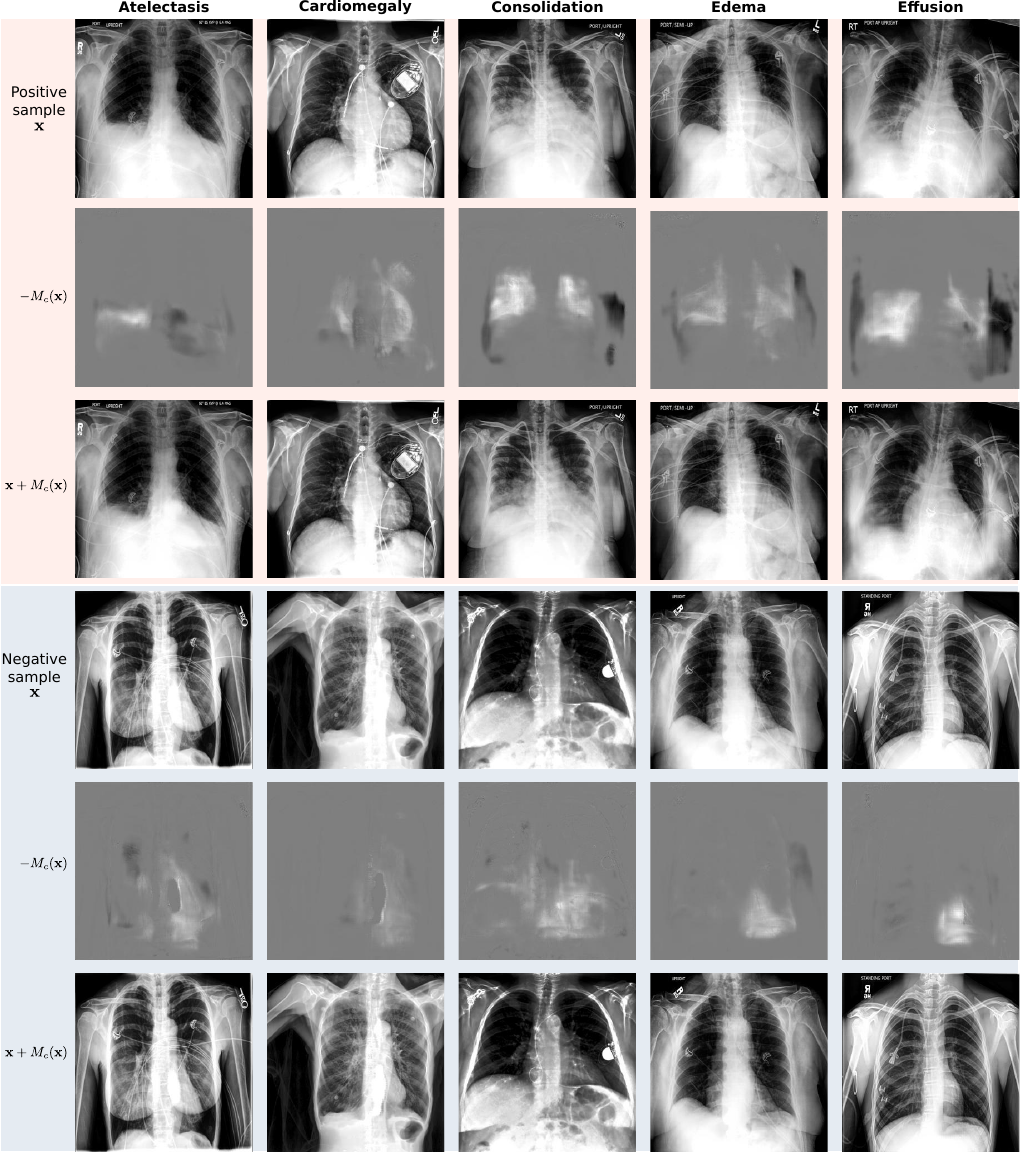}
    \caption{Counterfactual image generation. Samples from the CheXpert dataset. The examples in the top group of rows (in red) show input images that are positive for different classes $c$. For those, the evidence for specific disease $c$ is strong, as shown in attribution maps $M_c(\mathbf{x})$. After adding $M_c(\mathbf{x})$ with inputs, the strong disease effects are removed, as shown in the counterfactuals in the third row. The bottom group of rows (in blue) shows images that are negative for class $c$. For these images, the disease effects in attribution maps are light, and those images remain mostly unchanged by adding the output of $M_c(\mathbf{x})$.}
    \label{fig:counterfactual_appendix}
\end{figure*}

Examples of counterfactual images obtained by adding the class-specific attribution map to the input image, i.e. $\hat{\mathbf{x}} = \mathbf{x} + M_c(\mathbf{x})$, are shown in Fig.~\ref{fig:counterfactual_appendix}. 

\section{Additional results of classification}
\label{sec:appendix_classification}

Table \ref{tab:auc_each_disease} shows the disease-wise classification performance of all compared models for reference.

\begin{table*}[b]
    \centering
    \caption{Classification performance for each disease, measured by area under the ROC curve (AUC).}
    \label{tab:auc_each_disease}
    \resizebox{\textwidth}{!}{%
    \begin{tabular}{c|cccccc}
        \hline
         &\multicolumn{6}{c}{\bfseries CheXpert}\\
         \textbf{Model} & \bfseries Atelectasis & \bfseries Cardio. & \bfseries Consolid. & \bfseries Edema & \bfseries Effusion & \bfseries Avg.\\
        \hline
        ResNet50 &0.782 &0.902 &0.867 &0.877 &0.950 &0.875\\
        B-cos ResNet50 &0.791 &0.909 &0.846 &0.855 &0.928 &0.866 \\
        B-cos ResNet50 (guided) &0.719 &0.904 &0.820 &0.840 &0.915 &0.839 \\
        ours &0.807 &0.914 &0.870 &0.841 &0.935 &0.873 \\
        ours (guided) &0.763 &0.852 &0.870 &0.839 &0.917 &0.848 \\
        \hline
        &\multicolumn{6}{c}{\bfseries ChestX-ray8}\\
         \textbf{Model} & \bfseries Atelectasis & \bfseries Cardio. & \bfseries Consolid. & \bfseries Edema & \bfseries Effusion & \bfseries Avg.\\
        \hline
        ResNet50 &0.723 &0.857 &0.708 &0.804 &0.799 &0.778 \\
        B-cos ResNet50 &0.697 &0.825 &0.694 &0.802 &0.766 & 0.757\\
        B-cos ResNet50 (guided) &0.692 &0.826 &0.696 &0.794 &0.760 &0.754 \\
        ours &0.713 &0.858 &0.711 &0.823 &0.792 &0.779 \\
        ours (guided) &0.718 &0.847 &0.699 &0.810 &0.794 &0.774 \\
        \hline
        &\multicolumn{6}{c}{\bfseries Vindr-CXR}\\
         \textbf{Model} & \bfseries Aortic enlarg. & \bfseries Cardio. & \bfseries Pulmon. fib. & \bfseries Pleu. thicken. & \bfseries Pleu. Effusion & \bfseries Avg.\\
        \hline
        ResNet50 &0.795 &0.827 &0.685 &0.720 &0.794 &0.764  \\
        B-cos ResNet50 &0.877 &0.942 &0.736 &0.781 &0.842 &0.836 \\
        B-cos ResNet50 (guided) &0.876 &0.926 &0.726 &0.773 &0.840 &0.828 \\
        ours &0.758 &0.872 &0.723 &0.753 &0.839 &0.789 \\
        ours (guided) &0.812 &0.860 &0.702 &0.743 &0.793 &0.782\\
        \hline
  \end{tabular}
  }
\end{table*}

\section{Ablation study of the model guidance}
\label{sec:ablation_guidance}

\begin{figure*}[h]
  \includegraphics[width=2\columnwidth]{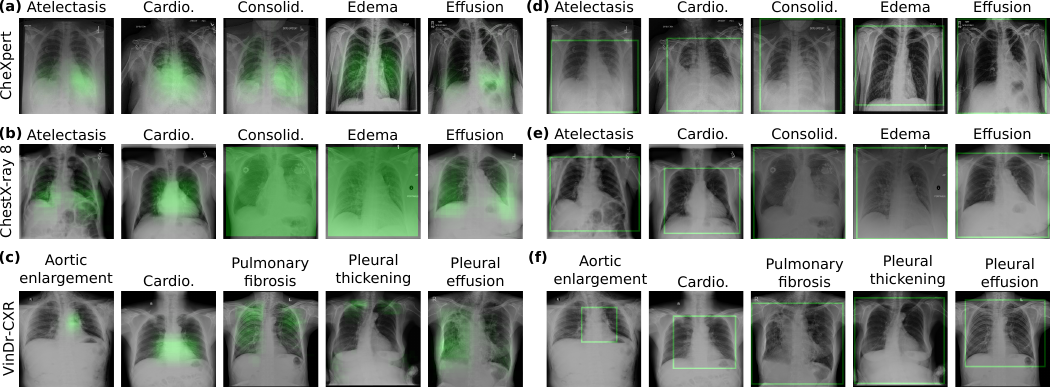}
  \caption{Weighted pseudo masks and pseudo bounding box.}
  \label{fig:all_pseudo_guidance}
\end{figure*}

We performed an ablation study of model guidance. As we discuss in the main paper, there are various approaches to generate pseudo guidance from the limited number of ground truth pixel-wise annotations of disease. We evaluated model guidance with the pseudo bounding box, pseudo-binary mask, and weighted pseudo mask and found that the pseudo binary mask perform best among the three kinds of pseudo guidance. We show the pseudo masks for three datasets and the result in the main paper. Here, we show the weighted pseudo masks (see Fig.~\ref{fig:all_pseudo_guidance} a, b, c)and pseudo bounding box (see Fig.~\ref{fig:all_pseudo_guidance} d, e, f) as an additional reference. It can be clearly observed that the weighted pseudo masks have very strong focuses while the pseudo bounding boxes cover much larger regions.

\begin{figure*}[h]
    \centering
    \includegraphics[width=2\columnwidth]{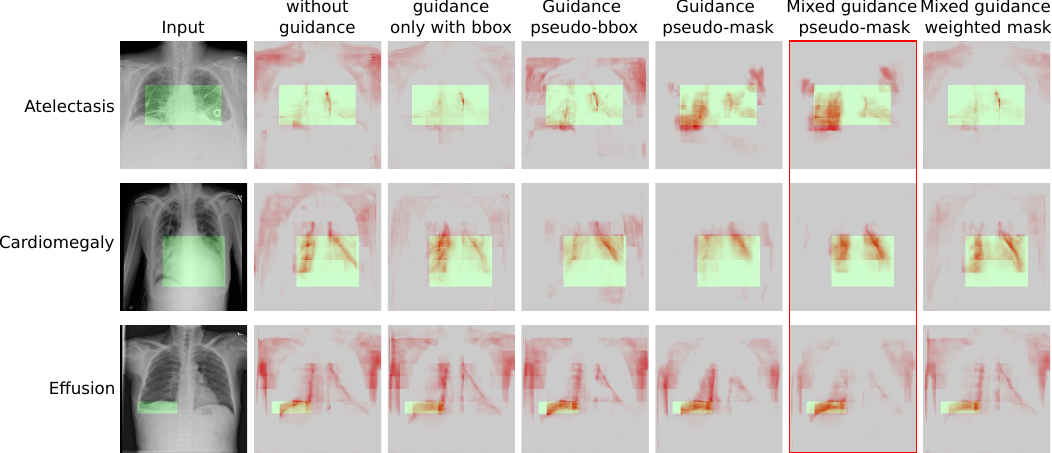}
    \caption{Local explanations from models trained with different model guidance approaches. Samples are from ChestX-ray8 dataset. The red box highlights the explanations from the best-performed ``mixed" model that is trained with both ground truth annotation and pseudo binary masks. }
    \label{fig:abla_guidance}
\end{figure*}

With Table \ref{tab:abla_guidance} and Fig.~\ref{fig:abla_guidance}, we show the quantitative and qualitative results of using different model guidance strategies. We have the following observations.

First, the classification performance and class sensitivity do not change much with different model guidance approaches. However, adding guidance to the model significantly improved the disease sensitivity, indicating that the model focuses more on the correct disease-relevant region. 

Second, the disease sensitivity score and the model guidance effect vary a lot between different diseases. For diseases with large lesion regions, such as cardiomegaly, the disease sensitivity score and the improvement of the score after adding guidance are much larger than those with smaller lesion regions, such as atelectasis.

Third, the model guidance approach affects disease localization a lot. The ``box only'' model is guided only using very limited pixel-wise annotations. The 30\% improvement in disease sensitivity of the ``box only'' model compared with the model ``without guidance'' shows that even very few annotations can achieve nice guidance. Since chest X-ray images have relative fixed structures, the model ``pseudo bbox" trained only with pseudo bounding boxes achieved better performance than the model using only ground truth annotation. The improvement of the model ``pseudo mask" which is trained only with pseudo masks compared with model ``pseudo bbox" shows that more focused guidance have better effects. The ``mixed" model is trained using the strategy we describe in the main paper, i.e., for samples with ground truth annotations, use ground truth as guidance, otherwise, using pseudo masks as guidance. The best disease sensitive score achieved by the ``mixed" model leads us to the conclusion that we should make full use of the available limited ground truth annotations. Since the pseudo masks are not ground truth annotation, therefore, model ``mixed weighted mask" that guided by more focused weighted pseudo masks does not achieve the best performance in disease sensitivity. 

\clearpage

\begin{table*}[t]
    \centering
    \caption{Ablation study on model guidance. Evaluated on ChestX-ray8 dataset.}
    \label{tab:abla_guidance}  
    \resizebox{\textwidth}{!}{%
    \begin{tabular}{c|c|c|c|ccc}
        \hline
        \bfseries Guidance & \bfseries Classification &\bfseries Class  &\bfseries Disease sensitivity
        &\bfseries &\bfseries Disease sensitivity &\bfseries\\  
        
        & \bfseries AUC & \bfseries sensitivity  & \bfseries average over disease &\bfseries Atelectasis &\bfseries Cardiomegaly &\bfseries Effusion\\
        \hline
        without guidance &{0.779} & 0.951 &0.158  & 0.047 & 0.343 & 0.084\\
        bbox only &\textbf{0.782} & 0.942 &0.201 &0.067 & 0.447 & 0.087\\
        pseudo bbox & 0.775 &0.950 &0.249 &0.069 &0.582 &0.097\\
        pseudo mask & 0.775 &0.944 &0.313 &\textbf{0.139} &0.689 &0.110\\
        mixed &0.774 & \textbf{0.965} &\textbf{0.327} &0.120 & {0.733} & \textbf{0.129}\\
        mixed weighted mask &0.780 & 0.948 &0.305 & 0.071 & \textbf{0.753} & 0.092\\
        \hline      
  \end{tabular}
  }
  \vspace{-0.3cm}
\end{table*}

\clearpage
\section{Additional qualitative results of local explanation} 
\label{sec:qualitative_results}

\begin{figure*}[h]
    \centering
    \includegraphics[width=1.8\columnwidth]{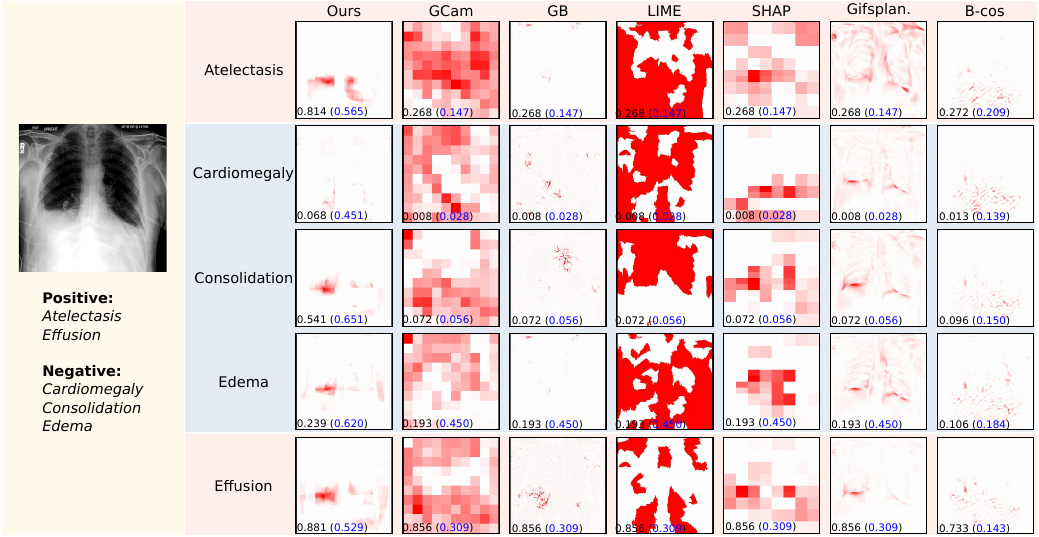}
    \caption{Local explanations for an example image from the CheXpert dataset.}
    \label{fig:local_exps_chexpert}
\end{figure*}

\begin{figure*}[h]
    \centering
    \includegraphics[width=1.8\columnwidth]{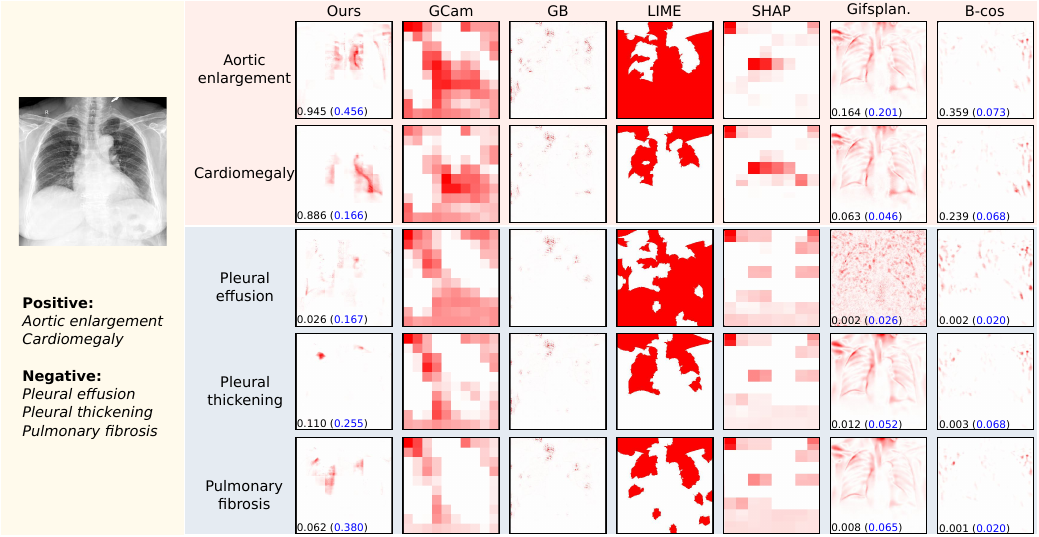}
    \caption{Local explanations for an example image from the Vindr-CXR dataset.}
    \label{fig:local_exps_vindr}
\end{figure*}

Fig.~\ref{fig:local_exps_chexpert} and Fig.~\ref{fig:local_exps_vindr} contain additional examples of local explanations using all compared methods derived from the CheXpert and Vindr-CXR datasets, respectively.

\section{Additional quantitative results of local explanation. } 
\label{sec:quantitative_results}

We present disease-specific quantitative results, including 95\% confidence intervals for class sensitivity and disease sensitivity, across all three datasets in the Tables \ref{tab:sup_class_sensitivity} and \ref{tab:sup_disease_sensitivity} for further reference. 

As discussed in Appendix~\ref{sec:ablation_guidance}, disease sensitivity scores vary greatly between different diseases. On three datasets, the disease sensitivity scores of cardiomegaly are much higher than those scores of other diseases. The improvement of disease sensitivity in cardiomegaly after adding guidance is also much bigger. This can be explained by the relatively large and fixed lesion region of the disease cardiomegaly. The big improvement after adding model guidance also happens with disease arotic enlargement, which has a quite focused lesion region which can be seen from Fig.~\ref{fig:abla_guidance}. Notably, on three datasets, the disease sensitivity scores of pleural effusion do not improve much after adding guidance. By comparing the weighted pseudo-masks of effusion on three datasets in Fig.~\ref{fig:abla_guidance}, it is clear that the lesion regions of effusion vary much between different datasets. Therefore, the pseudo guidance mechanism does not perform well in this disease.

\section{Implementation details}
\label{sec:training}

\subsection{Discriminator training}
\label{sec:discriminator_training}

The Attri-Net framework requires training a discriminator function $D$ in parallel to the class attribution generator $M$. The weight parameters $\theta$ of the discriminator are computed in separate gradient update steps using the Wasserstein GAN objective. The full discriminator optimisation objective is then given by
$$ \min_\theta \sum_c \E_{\mathbf{x} \sim p(\mathbf{x}|y_c=0)}[D_{c}(\mathbf{x}|\theta)] + \E_{\mathbf{x} \sim p(\mathbf{x}|y_c=1)}[D_{c}(\mathbf{x} + M_c(\mathbf{x})|\theta)] \text{\,,}$$
where we omitted the gradient penalty loss which ensures the discriminator fulfills the Lipschitz-1 constraint dictated by the Wasserstein GAN objective.

\subsection{Network architectures}
\label{sec:network_architectures}
    
The network architecture of the attribution map generator and the discriminator of the Attri-Net framework are shown in Table \ref{tab:arch_generator} and Table. \ref{tab:arch_critic}, respectively. L refers to the length of input/output features, N is the number of output channels, and K is the kernel size. 

\clearpage

\begin{table*}[htbp]
    \centering
    \caption{Comparison of Class Sensitivity.}
    \label{tab:sup_class_sensitivity}
    \resizebox{0.95\textwidth}{!}{%
    \begin{tabular}{c|ccccc}
        \hline
         &\multicolumn{5}{c}{\bfseries CheXpert}\\
         \textbf{Model} & \bfseries Atelectasis & \bfseries Cardio. & \bfseries Consolid. & \bfseries Edema & \bfseries Effusion\\
         \hline
         GB &$0.267\pm0.197$ &$0.376\pm0.213$ &$0.329\pm0.211$ &$0.262\pm0.126$ &$0.293\pm0.249$\\
         GCam &$0.177\pm0.198$ &$0.189\pm0.209$ &$0.245\pm0.213$ &$0.182\pm0.144$ &$0.160\pm0.216$ \\
         LIME &$0.248\pm0.147$ &$0.254\pm0.132$ &$0.263\pm0.046$ &$0.244\pm0.049$ &$0.256\pm0.165$ \\
         SHAP &$0.290\pm0.160$ &$0.325\pm0.162$ &$0.319\pm0.116$ &$0.382\pm0.132$ &$0.431\pm0.181$ \\
         Gifsp. &$0.259\pm0.273$ &$0.497\pm0.593$ &$0.461\pm0.390$ &$0.167\pm0.271$ &$0.108\pm0.269$ \\
         B-cos ResNet50 &$0.230\pm0.200$ &$0.360\pm0.207$ &$0.211\pm0.091$ &$0.230\pm0.103$ &$0.294\pm0.140$ \\
         B-cos ResNet50 (guided)&$0.223\pm0.134$ &$0.320\pm0.115$ &$0.286\pm0.123$ &$0.249\pm0.110$ &$0.286\pm0.204$ \\
         ours &$\textbf{0.533}\pm0.353$ &$\textbf{0.743}\pm0.320$ &$\textbf{0.663}\pm0.239$ &$\textbf{0.715}\pm0.266$ &$\textbf{0.797}\pm0.226$ \\
         ours (pseudo guidance) &$0.499\pm0.186$ &$0.659\pm0.241$ &$0.533\pm0.270$ &$0.588\pm0.230$ &$0.796\pm0.238$ \\
         ours (full guidance) &- &- &- &- &- \\
         \hline
        &\multicolumn{5}{c}{\bfseries ChestX-ray8}\\
         \textbf{Model} & \bfseries Atelectasis & \bfseries Cardio. & \bfseries Consolid. & \bfseries Edema & \bfseries Effusion \\
        \hline
         GB &$0.302\pm0.123$ &$0.266\pm0.129$ &$0.228\pm0.086$ &$0.310\pm0.100$ &$0.205\pm0.088$\\
         GCam &$0.205\pm0.127$ &$0.226\pm0.103$ &$0.232\pm0.145$ &$0.219\pm0.136$ &$0.249\pm0.143$ \\
         LIME &$0.231\pm0.065$ &$0.207\pm0.075$ &$0.227\pm0.059$ &$0.247\pm0.113$ &$0.226\pm0.075$ \\
         SHAP &$0.453\pm0.078$ &$0.483\pm0.083$ &$0.395\pm0.097$ &$0.367\pm0.104$ &$0.468\pm0.095$ \\
         Gifsp. &$0.511\pm0.555$ &$0.830\pm0.549$ &$0.696\pm0.608$ &$0.785\pm0.585$ &$0.520\pm0.690$ \\
         B-cos ResNet50 &$0.308\pm0.163$ &$0.263\pm0.147$ &$0.280\pm0.140$ &$0.253\pm0.144$ &$0.277\pm0.158$ \\
         B-cos ResNet50 (guided)&$0.325\pm0.202$ &$0.282\pm0.204$ &$0.362\pm0.189$ &$0.360\pm0.182$ &$0.278\pm0.162$ \\
         ours &$0.925\pm0.045$ &$0.959\pm0.024$ &$0.944\pm0.024$ &$0.954\pm0.023$ &$\textbf{0.972}\pm0.028$ \\
         ours (pseudo guidance) &$\textbf{0.954}\pm0.021$ &$\textbf{0.971}\pm0.019$ &$\textbf{0.958}\pm0.029$ &$\textbf{0.972}\pm0.020$ &$0.968\pm0.013$ \\
         ours (full guidance) &- &- &- &- &- \\
         \hline
        &\multicolumn{5}{c}{\bfseries Vindr-CXR}\\
         \textbf{Model} & \bfseries Aortic enlarg. & \bfseries Cardio. & \bfseries Pulmon. fib. & \bfseries Pleu. thicken. & \bfseries Pleu. Effusion \\
          \hline
         GB &$0.236\pm0.140$ &$0.250\pm0.135$ &$0.274\pm0.112$ &$0.296\pm0.139$ &$0.268\pm0.139$\\
         GCam &$0.175\pm0.105$ &$0.174\pm0.126$ &$0.160\pm0.140$ &$0.152\pm0.132$ &$0.219\pm0.074$ \\
         LIME &$0.255\pm0.047$ &$0.230\pm0.144$ &$0.267\pm0.087$ &$0.258\pm0.106$ &$0.245\pm0.063$ \\
         SHAP &$0.362\pm0.137$ &$0.350\pm0.106$ &$0.254\pm0.074$ &$0.270\pm0.091$ &$0.288\pm0.083$ \\
         Gifsp. &$0.217\pm0.502$ &$0.170\pm0.444$ &$0.297\pm0.411$ &$0.270\pm0.398$ &$0.672\pm0.650$ \\
         B-cos ResNet50 &$0.251\pm0.137$ &$0.271\pm0.145$ &$0.217\pm0.122$ &$0.227\pm0.118$ &$0.242\pm0.122$ \\
         B-cos ResNet50 (guided)&$0.252\pm0.154$ &$0.230\pm0.117$ &$0.241\pm0.147$ &$0.218\pm0.142$ &$0.248\pm0.103$ \\
         ours &$0.885\pm0.130$ &$0.883\pm0.146$ &$0.850\pm0.210$ &$0.866\pm0.170$ &$0.821\pm0.290$ \\
         ours (pseudo guidance) &$\textbf{0.938}\pm0.079$ &$0.912\pm0.145$ &$\textbf{0.918}\pm0.079$ &$\textbf{0.908}\pm0.109$ &$\textbf{0.933}\pm0.087$ \\
         ours (full guidance) &$0.885\pm0.132$ &$\textbf{0.914}\pm0.114$ &$0.818\pm0.258$ &$0.889\pm0.088$ &$0.856\pm0.195$ \\
        \hline
    \end{tabular}
    }
\end{table*}

\begin{table*}[htbp]
    \centering
    \caption{Comparison of Disease Sensitivity.}
    \label{tab:sup_disease_sensitivity}
    \resizebox{0.95\textwidth}{!}{%
    \begin{tabular}{c|ccccc}
        \hline
         &\multicolumn{5}{c}{\bfseries CheXpert}\\
         \textbf{Model} & \bfseries Atelectasis & \bfseries Cardio. & \bfseries Consolid. & \bfseries Edema & \bfseries Effusion\\
         \hline
         GB &$0.056\pm0.071$ &$0.255\pm0.176$ &$0.207\pm0.297$ &$0.210\pm0.224$ &$0.149\pm0.229$\\
         GCam &$0.102\pm0.135$ &$0.343\pm0.371$ &$0.141\pm0.212$ &$0.264\pm0.240$ &$0.120\pm0.237$ \\
         LIME &$0.115\pm0.128$ &$0.099\pm0.088$ &$0.032\pm0.078$ &$0.163\pm0.154$ &$0.102\pm0.190$ \\
         SHAP &$0.115\pm0.139$ &$0.236\pm0.244$ &$0.177\pm0.243$ &$0.404\pm0.284$ &$0.165\pm0.317$ \\
         Gifsp. &$0.148\pm0.180$ &$0.190\pm0.163$ &$0.175\pm0.190$ &$0.301\pm0.213$ &$0.144\pm0.240$ \\
         B-cos ResNet50 &$0.187\pm0.221$ &$0.414\pm0.238$ &$0.211\pm0.230$ &$0.273\pm0.243$ &$0.211\pm0.331$ \\
         B-cos ResNet50 (guided)&$0.232\pm0.251$ &$0.480\pm0.240$ &$0.226\pm0.209$ &$0.239\pm0.257$ &$\textbf{0.217}\pm0.341$ \\
         ours &$0.199\pm0.341$ &$0.223\pm0.248$ &$0.165\pm0.225$ &$0.353\pm0.277$ &$0.093\pm0.246$ \\
         ours (pseudo guidance) &$\textbf{0.374} \pm 0.451$  & $\textbf{0.623} \pm 0.356$ & $\textbf{0.324} \pm 0.414$ & $\textbf{0.490} \pm 0.329$ & $0.194\pm 0.418 $ \\
         ours (full guidance) &- &- &- &- &- \\
         \hline
        &\multicolumn{5}{c}{\bfseries ChestX-ray8}\\
         \textbf{Model} & \bfseries Atelectasis & \bfseries Cardio. & \bfseries Consolid. & \bfseries Edema & \bfseries Effusion \\
        \hline
         GB &$0.093\pm0.205$ &$0.345\pm0.231$ &- &- &$0.089\pm0.169$\\
         GCam &$0.052\pm0.141$ &$0.248\pm0.157$ &- &- &$0.077\pm0.154$ \\
         LIME &$0.049\pm0.116$ &$0.237\pm0.118$ &- &- &$0.080\pm0.132$ \\
         SHAP &$0.098\pm0.217$ &$0.598\pm0.255$ &- &- &$\textbf{0.139}\pm0.247$ \\
         Gifsp. &$0.084\pm0.191$ &$0.319\pm0.239$ &- &- &$0.121\pm0.187$ \\
         B-cos ResNet50 &$0.098\pm0.227$ &$0.480\pm0.243$ &- &- &$0.127\pm0.208$ \\
         B-cos ResNet50 (guided)&$0.098\pm0.244$ &$0.518\pm0.215$ &- &- &$0.125\pm0.223$ \\
         ours &$0.047\pm 0.112$ &0.343$\pm 0.251$ &- &- &$0.084\pm 0.161$ \\
         ours (pseudo guidance) &$\textbf{0.120}\pm0.260$ &$\textbf{0.733}\pm0.351$ &- &- &$0.129\pm0.257$ \\
         ours (full guidance) &- &- & - & - & - \\
         \hline
        &\multicolumn{5}{c}{\bfseries Vindr-CXR}\\
         \textbf{Model} & \bfseries Aortic enlarg. & \bfseries Cardio. & \bfseries Pulmon. fib. & \bfseries Pleu. thicken. & \bfseries Pleu. Effusion \\
          \hline
         GB &$0.041\pm0.097$ &$0.089\pm0.161$ &$0.026\pm0.089$ &$0.014\pm0.072$ &$0.065\pm0.160$\\
         GCam &$0.036\pm0.075$ &$0.221\pm0.179$ &$0.017\pm0.054$ &$0.006\pm0.020$ &$0.028\pm0.090$ \\
         LIME &$0.050\pm0.052$ &$0.034\pm0.077$ &$0.022\pm0.053$ &$0.007\pm0.024$ &$0.039\pm0.091$ \\
         SHAP &$0.105\pm0.160$ &$0.168\pm0.150$ &$0.006\pm0.022$ &$0.005\pm0.016$ &$0.050\pm0.067$ \\
         Gifsp. &$0.071\pm0.090$ &$0.127\pm0.113$ &$0.024\pm0.057$ &$0.010\pm0.022$ &$0.029\pm0.077$ \\
         B-cos ResNet50 &$0.119\pm0.163$ &$0.250\pm0.196$ &$0.012\pm0.048$ &$0.009\pm0.030$ &$0.056\pm0.133$ \\
         B-cos ResNet50 (guided)&$0.120\pm0.181$ &$0.190\pm0.191$ &$0.012\pm0.055$ &$0.012\pm0.032$ &$0.042\pm0.115$ \\
         ours &$0.119\pm 0.180$ &0.196$\pm 0.164$ &0.017$\pm0.049$ &$0.010\pm0.030$ &0.034$\pm0.129$ \\
        ours (pseudo guidance) &0.270$\pm0.433$ &$0.345\pm0.314$ &\textbf{0.053}$\pm0.178$ &\textbf{0.035}$\pm0.127$ &0.077$\pm0.231$ \\
         ours (full guidance) &\textbf{0.363}$\pm0.560$ &\textbf{0.504}$\pm0.466$ &$0.031\pm0.115$ &$0.022\pm0.056$ &$\textbf{0.102}\pm0.282$ \\
        \hline
    \end{tabular}
    }
\end{table*}

\begin{table*}[h]
  \centering
  \caption{Attri-Net class attribution generator network architecture.}%
  \label{tab:arch_generator}
  \resizebox{\textwidth}{!}{%
\begin{tabular}{lcc}
\hline
\bfseries Layers     & \bfseries Input $\rightarrow$ Output       & \bfseries Layer information \\
\hline
\bfseries Task embedding layer & Task code $\mathbf{t}_c$ $\rightarrow$ Task embedding $\mathbf{t}_c^\prime$   & 8 $\times$ FC(L100,L100)\\
\hline 

                      &    & Ada\_Conv: CONV(N64, K7x7), AdaIN, ReLU\\
\bfseries Down-sampling & (Input image $\mathbf{x}$, $\mathbf{t}_c^\prime$) $\rightarrow$ $\mathbf{out}_{\text{down}}$       & Ada\_Conv: CONV(N128, K4x4), AdaIN, ReLU \\
                      &    & Ada\_Conv: CONV(N256, K4x4), AdaIN, ReLU \\
\hline
                      &    & Ada\_ResBlock: CONV(N256, K3x3), AdaIN, ReLU \\
                      &     & Ada\_ResBlock: CONV(N256, K3x3), AdaIN, ReLU \\
\bfseries Bottlenecks  & ($\mathbf{out}_{\text{down}}$ , $\mathbf{t}_c^\prime$) $\rightarrow$ $\mathbf{out}_{\text{bn}}$  & Ada\_ResBlock: CONV(N256, K3x3,), AdaIN, ReLU \\
                     &     & Ada\_ResBlock: CONV(N256, K3x3), AdaIN, ReLU \\
                     &     & Ada\_ResBlock: CONV(N256, K3x3), AdaIN, ReLU \\
                    &     & Ada\_ResBlock: CONV(N256, K3x3), AdaIN, ReLU \\
  \hline
                        & & Ada\_DECONV(N128, K4x4), AdaIN, ReLU \\
\bfseries Up-sampling & ($\mathbf{out}_{\text{bn}}$ , $\mathbf{t}_c^\prime$) $\rightarrow$ $\mathbf{out}_{\text{up}}$ & Ada\_DECONV(N64, K4x4), AdaIN, ReLU \\
          & & CONV(N1, K7x7) \\
  \hline
\bfseries Output layer & ($\mathbf{x}$, $\mathbf{out}_{\text{up}}$) $\rightarrow$  $M_c(\mathbf{x})$ &  $M_c(\mathbf{x}) = \text{tanh}(\mathbf{x}+\mathbf{out}_{\text{up}})-\mathbf{x}$\\
  \hline
    \end{tabular}
  }
\end{table*}

\begin{table*}[h]
  \centering
  \caption{Attri-Net discriminator network architecture.}%
  \label{tab:arch_critic}
  \vspace{1mm}
  \resizebox{\textwidth}{!}{%
\begin{tabular}{lcc}
\hline
\bfseries Layers     & \bfseries Input $\rightarrow$ Output       & \bfseries Layer information \\
\hline
\bfseries Task embedding layer & Task code $\mathbf{t}_c$ $\rightarrow$ Task embedding $\mathbf{t}_c^\prime$   & 8 $\times$ FC(L100,L100)\\
\hline 
\bfseries Input layer & & Ada\_Conv: CONV(N64, K4x4), AdaIN, ReLU \\
& & Ada\_Conv: CONV(N128, K4x4), AdaIN, ReLU \\
& & Ada\_Conv: CONV(N256, K4x4), AdaIN, ReLU \\
\bfseries Hidden layers & ($\mathbf{x} \slash \hat{\mathbf{x}}$ , $\mathbf{t}_c^\prime$) $\rightarrow$ $\mathbf{out}_{\text{hid}}$  & Ada\_Conv: CONV(N512, K4x4), AdaIN, ReLU \\
& & Ada\_Conv: CONV(N1024, K4x4), AdaIN, ReLU \\
& & Ada\_Conv: CONV(N2048, K4x4), AdaIN, ReLU \\
  \hline
\bfseries Output layer  & $\mathbf{out}_{\text{hid}}$  $\rightarrow$ $\mathcal{L}^{(c)}_{\text{adv}}$ & CONV(N1, K3x3)\\
  \hline
    \end{tabular}
  }
\end{table*}

\end{document}